\documentclass{article}

\usepackage{hyperref}

\usepackage{fontawesome5}
\usepackage{xcolor}
\newcommand{\hflogo}{\raisebox{-0.2em}{\includegraphics[height=1em]{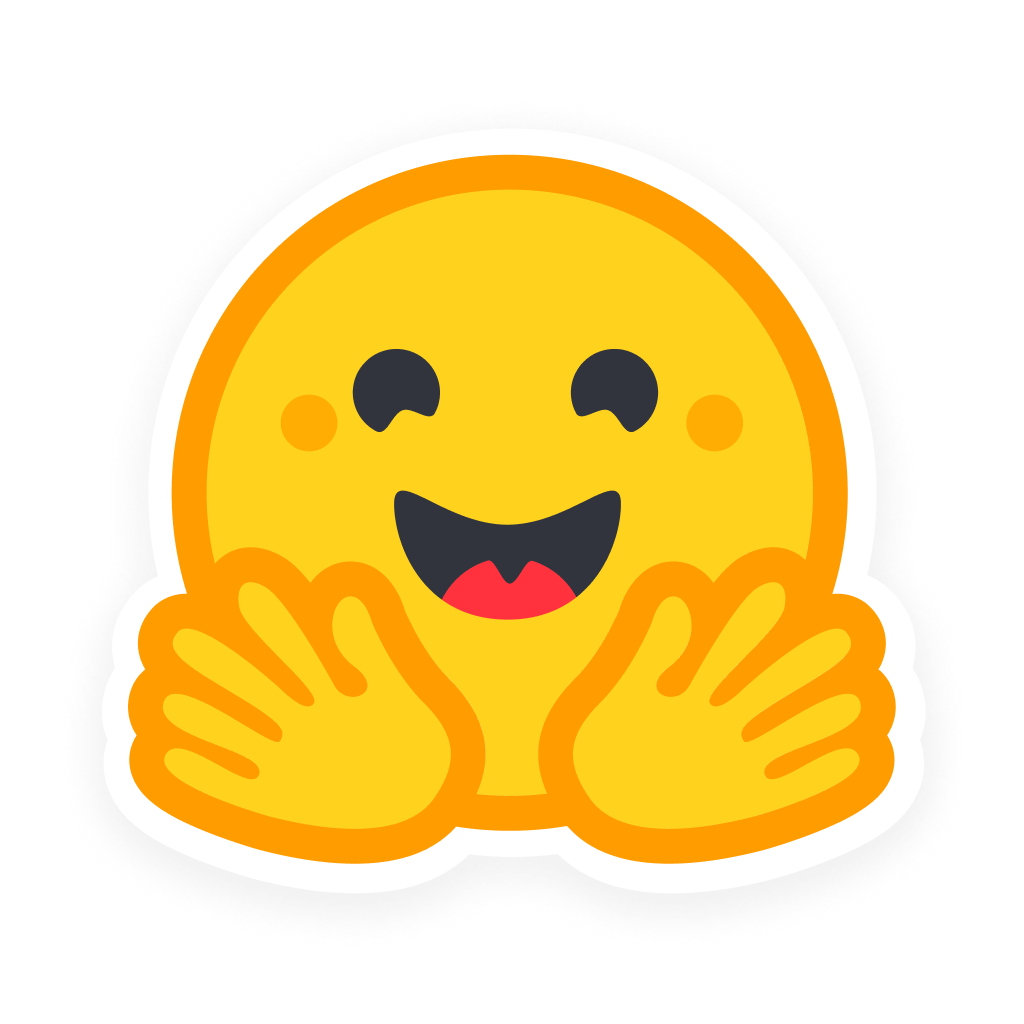}}}

\definecolor{githubcolor}{HTML}{333333}
\definecolor{hfcolor}{HTML}{FFD21E}
\definecolor{webcolor}{HTML}{4285F4}

\usepackage[final]{neurips_2025}

\usepackage{float}
\usepackage[utf8]{inputenc} 
\usepackage[T1]{fontenc}    
\usepackage{hyperref}       
\usepackage{url}            
\usepackage{booktabs}       
\usepackage{amsfonts}       
\usepackage{nicefrac}       
\usepackage{microtype}      
\usepackage{subcaption}
\usepackage{amsmath} 
\usepackage{amssymb}
\usepackage{mathtools}
\usepackage{amsthm}
\usepackage{graphicx}
\usepackage{booktabs} 
\newcommand{\thickmidrule}{\specialrule{1.5pt}{0pt}{0pt}} 
\usepackage{paralist}
\usepackage[compact]{titlesec}
\usepackage[capitalize,noabbrev]{cleveref}
 
\theoremstyle{plain}

\theoremstyle{definition}

\theoremstyle{remark}

\usepackage{graphicx} 
\usepackage{caption} 
\usepackage{enumitem} 
\usepackage{times} 
\usepackage{setspace} 
\usepackage{cite} 
\usepackage{hyperref}
\usepackage{multirow}
\usepackage{amssymb}
\usepackage[table]{xcolor}
\usepackage{booktabs}
\usepackage{rotating}
\usepackage{caption}
\usepackage{colortbl}
\definecolor{Gray}{gray}{0.90}
\newcolumntype{g}{>{\columncolor{Gray}}c}
\definecolor{ffe1da}{RGB}{255,225,218}
\definecolor{F7E0D5}{RGB}{247,224,213}
\definecolor{lightOrang}{RGB}{247,224,213}
\definecolor{darkF7E0D5}{RGB}{209,154,128}
\definecolor{FFE5CC}{RGB}{255,229,204}
\colorlet{Light}{FFE5CC}

\usepackage[textsize=tiny]{todonotes}
\usepackage{tablefootnote}
\usepackage{threeparttable}

\title{seq-JEPA: Autoregressive Predictive Learning of Invariant-Equivariant World Models}

\author{%
  Hafez Ghaemi$^{1,2,3}$\thanks{Correspondence to \url{hafez.ghaemi@umontreal.ca}}\qquad Eilif B. Muller$^{1,2,3}$\thanks{Equal Contribution}\qquad Shahab Bakhtiari$^{1,2}$\footnotemark[2]\\[0.5em]
  $^1$Université de Montréal, $^2$Mila - Quebec AI Institute, $^3$CHU Sainte-Justine
}

\begin{document}

\maketitle

\begin{abstract}
  Joint-embedding self-supervised learning (SSL) commonly relies on transformations such as data augmentation and masking to learn visual representations, a task achieved by enforcing invariance or equivariance with respect to these transformations applied to two views of an image. This dominant two-view paradigm in SSL often limits the flexibility of learned representations for downstream adaptation by creating performance trade-offs between high-level invariance-demanding tasks such as image classification and more fine-grained equivariance-related tasks. In this work, we propose \emph{seq-JEPA}, a world modeling framework that introduces architectural inductive biases into joint-embedding predictive architectures to resolve this trade-off. Without relying on dual equivariance predictors or loss terms, seq-JEPA simultaneously learns two architecturally separate representations for equivariance- and invariance-demanding tasks. To do so, our model processes short sequences of different views (observations) of inputs. Each encoded view is concatenated with an embedding of the relative transformation (action) that produces the next observation in the sequence. These view-action pairs are passed through a transformer encoder that outputs an aggregate representation. A predictor head then conditions this aggregate representation on the upcoming action to predict the representation of the next observation. Empirically, seq-JEPA demonstrates strong performance on both equivariance- and invariance-demanding downstream tasks without sacrificing one for the other. Furthermore, it excels at tasks that inherently require aggregating a sequence of observations, such as path integration across actions and predictive learning across eye movements.
\end{abstract}

\vspace{0.5em}
\begin{center}
\href{https://hafezgh.github.io/seq-jepa/}{\faGlobe~Project Page} \quad
\href{https://github.com/hafezgh/seq-jepa}{\faGithub~Code} \quad
\href{https://huggingface.co/Hafez/seq-JEPA}{\hflogo~Models} \quad
\href{https://huggingface.co/datasets/Hafez/salmap-stl10}{\hflogo~STL10 Saliency} \quad
\href{https://huggingface.co/datasets/Hafez/ImageNet1k-Saliency-Maps}{\hflogo~ImageNet-1k Saliency} \quad
\href{https://huggingface.co/datasets/Hafez/3DIEBench-OOD}{\hflogo~3DIEBench-OOD}
\end{center}

\section{Introduction}
\label{introduction}

Self-supervised learning (SSL) in latent space has made significant progress in visual representation learning, closing the gap with supervised methods across many tasks. Most SSL methods rely on comparing two transformed views of an image and enforcing invariance to the transformations~\citep{misra_self-supervised_2020,chen_simple_2020,he_momentum_2020,dwibedi_little_2021,haochen_provable_2021,yeh_decoupled_2022,caron_unsupervised_2020,caron_emerging_2021,ermolov_whitening_2021,assran_masked_2022,zbontar_barlow_2021,bardes_vicreg_2022}. Another group of methods employ techniques to preserve transformation-specific information, thereby learning equivariant representations \citep{lee_improving_2021,xiao_what_2021,park_learning_2022,dangovski_equivariant_2022,gupta_structuring_2023,garrido_self-supervised_2023,garrido_learning_2024,gupta_structuring_2023,gupta_-context_2024,yerxa_contrastive-equivariant_2024}.

Equivariance is a crucial representational property for downstream tasks that require fine-grained distinctions. For example, given representations that are invariant to color, it is not possible to distinguish between certain species of flowers or birds \citep{lee_improving_2021,xiao_what_2021}. Moreover, recent work has shown that equivariant representations are better aligned with neural responses in primate visual cortex and could be important for building more accurate models thereof \citep{yerxa_contrastive-equivariant_2024}. While some equivariant SSL approaches have reported minor gains on tasks typically associated with invariance (e.g., classification) \citep{devillers_equimod_2022,park_learning_2022,gupta_structuring_2023}, \emph{a growing body of work highlights a fundamental trade-off between learning invariance and equivariance, i.e., models that capture equivariance-related style latents do not fare well in classification and vice versa} \citep{garrido_self-supervised_2023, garrido_learning_2024, gupta_-context_2024,yerxa_contrastive-equivariant_2024,rusak_infonce_2025}. This trade-off has recently received theoretical support \citep{wang_understanding_2024}, underscoring the need for new architectural and objective designs that can reconcile these competing goals.

In contrast to the two-view paradigm in SSL, humans and other animals rely on a \emph{sequence} of actions and consequent observations (views) for developing appropriate visual representation during novel object learning ~\citep{harman_active_1999,vuilleumier_multiple_2002}. For example, they recognize a 3-D object by changing their viewpoint and examining different sides of the object~\citep{tarr_three-dimensional_1998}. Inspired by this, we introduce \textbf{seq-JEPA}, a self-supervised world modeling framework that combines joint-embedding predictive architectures~\citep{lecun_path_2022} with inductive biases for sequential processing. seq-JEPA simultaneously learns two architecturally distinct representations: one that is equivariant to a specified set of transformations, and another that is suited for invariance-demanding tasks, such as image classification.

Specifically, our framework (Figure~\ref{fig:schematic}) processes a short sequence of transformed views (observations) of an image. Each view is encoded and concatenated with an embedding corresponding to the relative transformation (action) that produces the next observation in the sequence. These view-action pairs are passed through a transformer encoder, \emph{a form of learned working memory}, that outputs an aggregate representation of them. A predictor head then conditions this aggregate representation on the upcoming action to predict the representation of the next observation.

Our results demonstrate that individual encoded views in seq-JEPA become transformation/action-equivariant. Through ablations, we show that action conditioning plays a key role in promoting equivariant representation learning in the encoder network. In contrast, the aggregate representation of views, produced at the output of the transformer proves highly effective for invariance-demanding downstream tasks. This emergent architectural disentanglement of invariance and equivariance is central to seq-JEPA's competitive performance compared to invariant and equivariant SSL methods on both categories of tasks (Figure~\ref{fig:inv-equi-plot}). Unlike most prior equivariant SSL methods \citep{lee_improving_2021,park_learning_2022,dangovski_equivariant_2022,gupta_structuring_2023,garrido_self-supervised_2023,gupta_-context_2024,yerxa_contrastive-equivariant_2024}, our model does not rely on explicitly crafted loss terms or objectives to achieve equivariance, nor is it instructed to learn the decomposition of two representations. Instead, the dual representation structure arises naturally from the model architecture and action-conditioned predictive learning.

Beyond resolving the invariance-equivariance trade-off, seq-JEPA further benefits from processing a sequence of observation views; we show that it performs well on tasks requiring integration over sequences of observations. In one scenario, inspired by embodied vision in primates, our model learns image representations without augmentations or masking, solely by predicting across simulated eye movements (saccades). In another setting, it performs path integration over sequences of actions—such as eye movements or 3D object rotations in 3DIEBench~\citep{garrido_self-supervised_2023}. \textbf{Our key contributions are as follows:}
\begin{compactitem}
\item We introduce seq-JEPA, a self-supervised world model that learns architecturally distinct representations for invariance- and equivariance-demanding downstream tasks through sequential prediction over action-observation pairs, without requiring explicit equivariance losses or dual predictors.
\item We empirically validate that seq-JEPA matches or outperforms existing invariant and equivariant SSL methods across tasks requiring either representational property.
\item We demonstrate that seq-JEPA naturally supports tasks that involve sequential integration of observations, such as predictive learning across saccades and path integration over action sequences.
\end{compactitem}

\section{Method}
\subsection{Invariant and equivariant representations}
Before presenting our architecture and training procedure, we briefly define invariance and equivariance in the context of SSL \citep{dangovski_equivariant_2022,devillers_equimod_2022}. Let $\mathcal{T}$ denote a distribution over transformations, parameterized by a vector $t$. These transformations—such as augmentations or masking—can be used to generate multiple views from a single image $x$. Let $x_1$ and $x_2$ be two such views, produced by applying transformations $t_1$ and $t_2$ sampled from $\mathcal{T}$. Additionally, let $a$ denote the \emph{relative} transformation that maps $x_1$ to $x_2$. Additionally, we denote $a$ as a transformation that transforms $x_1$ to $x_2$. We distinguish between $t$ (an individual transformation) and $a$ (an action), where the latter reflects the change from one view to another. Let $f$ be an encoder that maps inputs to a latent space. We say that $f$ is equivariant to $t$ if:

\begin{equation}
\label{eq:eq1}
\forall{t \in \mathcal{T}},~\exists{,u_t} ;\mathrm{s.t.} \qquad f(t(x)) = u_t(f(x)),
\end{equation}

where $u_t$ is a transformation in latent space corresponding to $t$. Equivariance can similarly be defined in terms of relative transformations (actions):

\begin{equation}
\label{eq:eq1a}
\forall{a \in \mathcal{T}},~ \exists{,u_a} ;\mathrm{s.t.} \qquad f(x_2) = u_a(f(x_1)).
\end{equation}

As a special case, if $u_t$ and $u_a$ are identity functions, then $f$ is invariant to the transformation: $f(t(x)) = f(x)$ or $f(x_2) = f(x_1)$.

\subsection{Architecture}
\begin{figure*}[t]
  \centering
  \begin{minipage}[t]{0.48\linewidth}
    \centering
    \includegraphics[width=\linewidth]{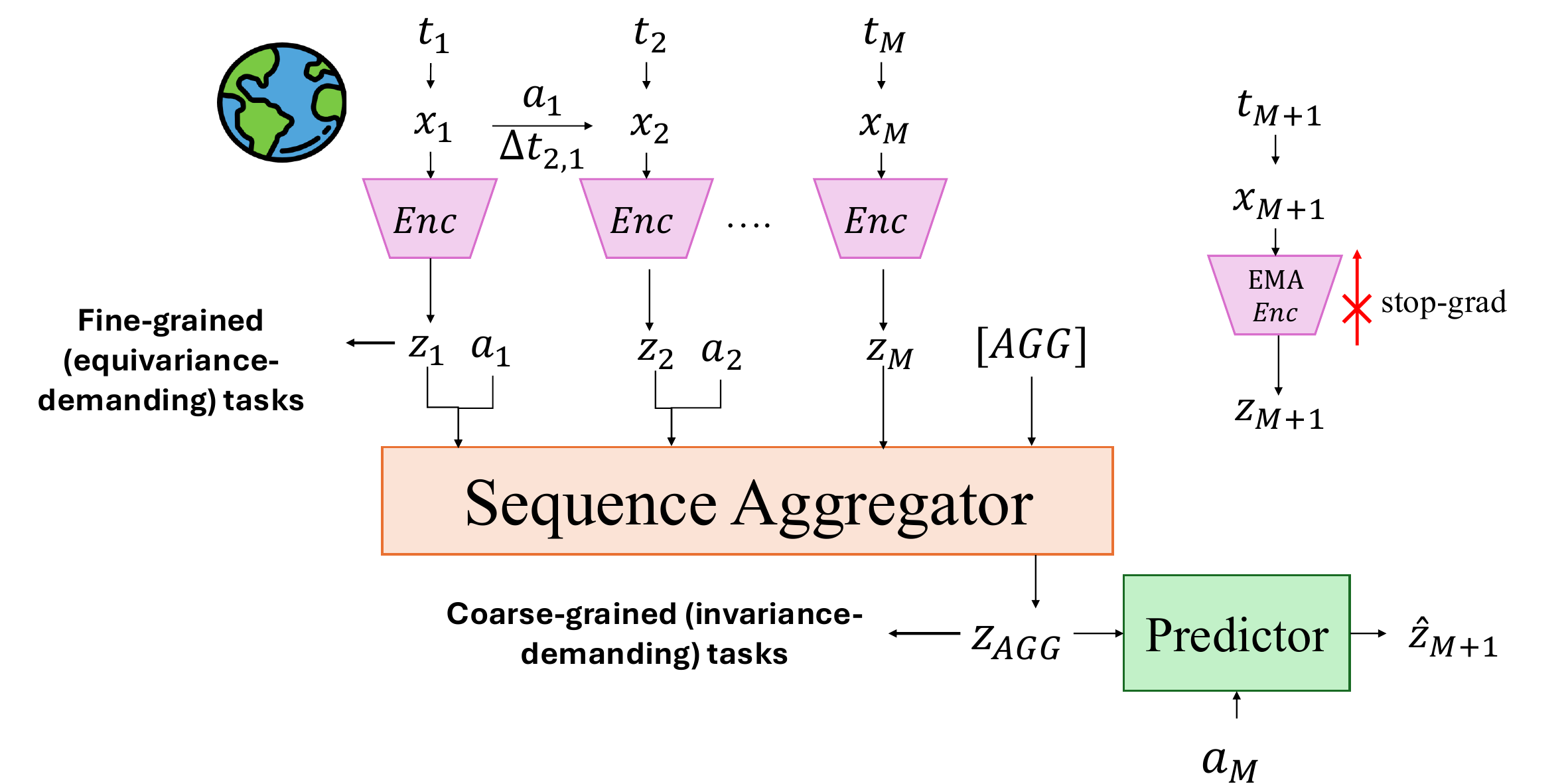}
    \captionof{figure}{\small seq-JEPA is a self-supervised world model that leverages a sequence of action-observation pairs to learn architecturally distinct representations for downstream tasks requiring transformation invariance or equivariance.}
    \label{fig:schematic}
  \end{minipage}
  \hfill
  \begin{minipage}[t]{0.48\linewidth}
    \centering
    \includegraphics[width=\linewidth]{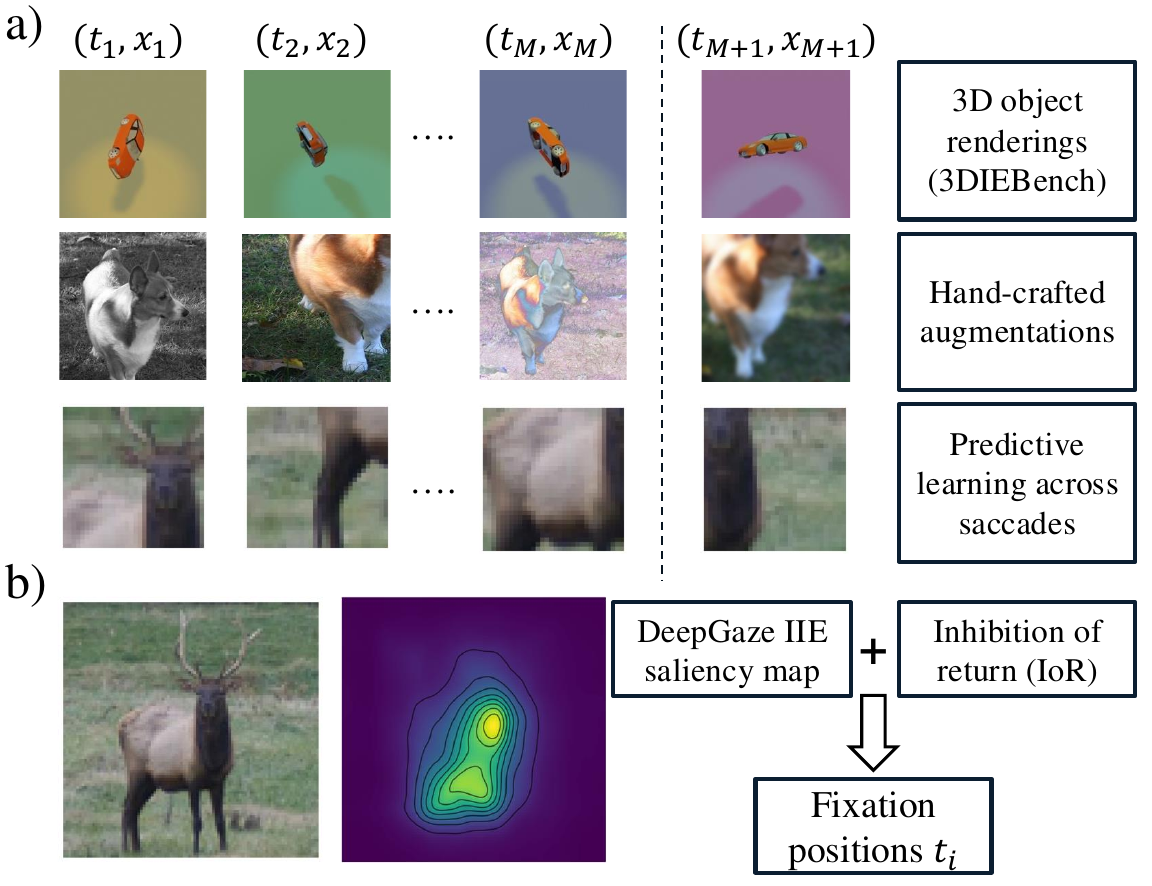}
    \captionof{figure}{\small \textbf{a.} Transformations and observations used for training \textbf{b.} In predictive learning across saccades, image saliencies and inhibition of return help create more informative and less redundant patch sequences.}
    \label{fig:setup}
  \end{minipage}
\end{figure*}

Figure~\ref{fig:schematic} presents the overall architecture of seq-JEPA. Let $\{x_i\}_{i=1}^{M+1}$ be a sequence of views generated from a sample $x$ via transformations $\{t_i\}_{i=1}^{M+1}$. The relative transformations (actions) $\{a_i\}_{i=1}^{M}$ are defined as $a_{i}\triangleq \Delta t_{i,i+1}$, i.e., the transformation mapping $x_i$ to $x_{i+1}$. In our default setting, we use a learnable linear projector to encode these actions.

A backbone encoder, $f$ encodes the first $M$ views, producing representations $\{z_i\}_{i=1}^{M}$. Except for $z_M$, each $z_i$ is concatenated with its corresponding action embedding and passed to a transformer encoder $g$ (no MLP projector is used after the encoder), which aggregates the sequence of action-observation pairs. The transformer uses a learnable \texttt{[AGG]} token (analogous to the \texttt{[CLS]} token in ViT~\citep{dosovitskiy_image_2020}) to generate the aggregate representation:

\begin{equation}
    z_{AGG}=g((z_1,a_1),(z_2,a_2)..., (z_{M-1},a_{M-1}), z_M)
\end{equation}

This aggregate representation $z_{AGG}$ is then concatenated with the final action embedding $a_M$ (corresponding to the transformation from $x_M$ to $x_{M+1}$), and passed to an MLP predictor $h$ to predict the representation of $x_{M+1}$:

\begin{equation}
\hat{z}_{M+1} = h(z_{AGG}, a_M).
\end{equation}

The ground truth $z_{M+1}$ is computed using a target encoder—an exponential moving average (EMA) of $f$. This target representation is passed through a stop-gradient operator (\texttt{sg}) to avoid representational collapse~\citep{grill_bootstrap_2020,chen_exploring_2021,assran_self-supervised_2023}. The training objective is to maximize the cosine similarity between $\hat{z}_{M+1}$ and $z_{M+1}$ with the loss function:

\newcommand{\lnorm}[1]{\frac{#1}{\left\lVert{#1}\right\rVert _2}}
\newcommand{\lnormv}[1]{{#1}/{\left\lVert{#1}\right\rVert _2}}
\begin{equation}
\mathcal{L}_\mathrm{seq-JEPA} = 1 - \lnorm{\hat{z}_{M+1}} \cdot \lnorm{\texttt{sg}(z_{M+1})}.
\end{equation}

No additional loss terms or equivariance-specific predictors are used during training.

\subsection{Action and observation sets}
To evaluate generalization across transformation types, we consider three sets of action-observation pairs (Figure~\ref{fig:setup}). See Appendix~\ref{sec:dataprep} for details of each setup.

\textbf{3D Invariant Equivariant Benchmark (3DIEBench).} The 3DIEBench dataset~\citep{garrido_self-supervised_2023} is designed to evaluate representational invariance and equivariance. It includes 3D object renderings with variations in rotation, floor hue, and lighting. In this benchmark, the action between two views is the relative difference of these three factors of variation. We primarily study equivariance to $SO(3)$ rotations and secondarily to appearance factors of floor and light hue.

\textbf{Hand-Crafted Augmentations.} In this setting, we use transformed views generated via common SSL augmentations (e.g., crop, color jitter, blur), and actions correspond to relative augmentation parameters. We use CIFAR100 and Tiny ImageNet for experiments in this setup, and follow EquiMod’s augmentation protocol~\citep{devillers_equimod_2022}.

\textbf{Predictive Learning Across Saccades (PLS).} Going beyond conventional transformations such as augmentations or 3D rotations, we show that seq-JEPA can learn visual representations from a sequence of partial observations thanks to architectural inductive biases—without relying on any hand-crafted augmentations. Our PLS has a similar flavor to I-JEPA \citep{assran_self-supervised_2023} but does not require engineered masking strategies. In PLS, we train seq-JEPA on sequences of patches extracted from full-resolution images. For instance, with STL-10 dataset, we use small $32 \times 32$ patches to form the observation sequence. In this setting, actions correspond to the relative positions between patch centers, simulating saccadic eye movements and inducing 2-D positional equivariance to representations. To select fixation points, we adopt two biologically inspired techniques that increase informativeness, reduce redundancy, and improve the downstream utility of the aggregate representation (Figure~\ref{fig:setup}):

\begin{compactitem}
    \item \textbf{Saliency-Based Fixation Sampling.} Using DeepGaze IIE~\citep{linardos_deepgaze_2021}, we extract saliency maps for each image and use them to probabilistically sample fixation points \citep{itti_model_1998,li_saliency_2002,zhaoping_v1_2014}. The maps are pre-computed and introduce no training overhead.
    \item \textbf{Inhibition of Return (IoR).} To reduce spatial overlap between patches and emulate natural exploration~\citep{posner_inhibition_1985}, we implement IoR by zeroing out the sampling probability of areas surrounding previously sampled fixations.
\end{compactitem}

\section{Experimental Setup}\label{sec:expsetup}
\subsection{Compared methods and baselines}

We compare seq-JEPA against both invariant and equivariant SSL baselines. Invariant methods include SimCLR~\citep{chen_simple_2020}, BYOL~\citep{grill_bootstrap_2020}, and VICReg~\citep{bardes_vicreg_2022} and VICReg with trajectory regularization~\citep{wang2024pose}. Equivariant methods include SEN~\citep{park_learning_2022}, EquiMod~\citep{devillers_equimod_2022}, SIE~\citep{garrido_self-supervised_2023}, and ContextSSL~\citep{gupta_-context_2024}. For all baselines, architectural details are given in Appendix~\ref{sec:archdetails}. We also evaluate two hybrid baselines based on our architecture:

\begin{compactitem}
\item \textbf{Conditional BYOL.} A two-view version of seq-JEPA with no sequence aggregator, where BYOL’s predictor is conditioned on the relative transformation between target and online views. This encourages representations to encode transformation information.
\item \textbf{Conv-JEPA.} A baseline for the saccades setting. It uses the same sequence of saliency-sampled patches as seq-JEPA and predicts the final patch’s representation from each earlier patch individually. These losses are summed across the pairs before backpropagation.
\end{compactitem}

\subsection{Training protocol} \label{sec:trprot}
All models use ResNet-18~\citep{he_deep_2016} as the backbone encoder. For action conditioning, we use a learnable linear projection to learn action embeddings (default action embedding is 128-d). The sequence aggregator in seq-JEPA is a lightweight transformer encoder \citep{vaswani_attention_2017} with three layers and four attention heads. The predictor is a 2-layer MLP with 1024 hidden units and ReLU activation. In order to control for and eliminate any performance gain resulting from using a transformer encoder in seq-JEPA instead of an MLP projection head, we trained baselines that typically use MLP projectors in two variants: (1) with original MLP projector; and (2) with the MLP replaced by a transformer encoder and a sequence length of one. We did not see any benefit from switching to transformer projectors in any of the baselines, and include the transformer-projector results in Appendix~\ref{sec:projresults}. All models are trained from scratch with a batch size of 512. We use 1000 epochs for 3DIEBench and 2000 epochs for other datasets to obtain asymptotic performance. We use AdamW for models with transformer projectors (including seq-JEPA) due to its stability and improved regularization in transformer training. For ConvNet-only models with MLP heads, we use the Adam optimizer. Full hyperparameters are detailed in Appendix~\ref{app1}.

\subsection{Evaluation metrics and protocol} \label{sec:valprot}
To assess equivariance, we follow the protocol of \citet{garrido_self-supervised_2023} and train  a regressor on frozen encoder representations to predict the relative transformation (action) between two views. We report the $R^2$ score on the test set. In addition to action decoding $R^2$, we also report retrieval-based metrics including Mean Reciprocal Rank (MRR), Hit@1, and Hit@5~\citep{kipf_contrastive_2020,park_learning_2022,garrido_self-supervised_2023} to evaluate the quality of the predictor. As a proxy measure of invariance, we use top-1 classification accuracy of a linear probe on top of frozen representations. For all baselines, probes are trained on encoder outputs. For seq-JEPA, we measure accuracies on top of the aggregate representation ($z_{AGG}$ in Figure~\ref{fig:schematic}) and report the number of observation views used during training and inference. For completeness, we also report classification performance on top of encoder representation ($z_i$ in Figure~\ref{fig:schematic}) for seq-JEPA models in Appendix~\ref{sec:resaccs}. Training details of evaluation heads are given in Appendix~\ref{sec:evaldetails}.

\section{Results}

\subsection{Quantitative evaluation on 3DIEBench}

\begin{figure*}[t]
  \centering
  \begin{minipage}{0.44\linewidth}
    \centering
    \includegraphics[width=\linewidth]{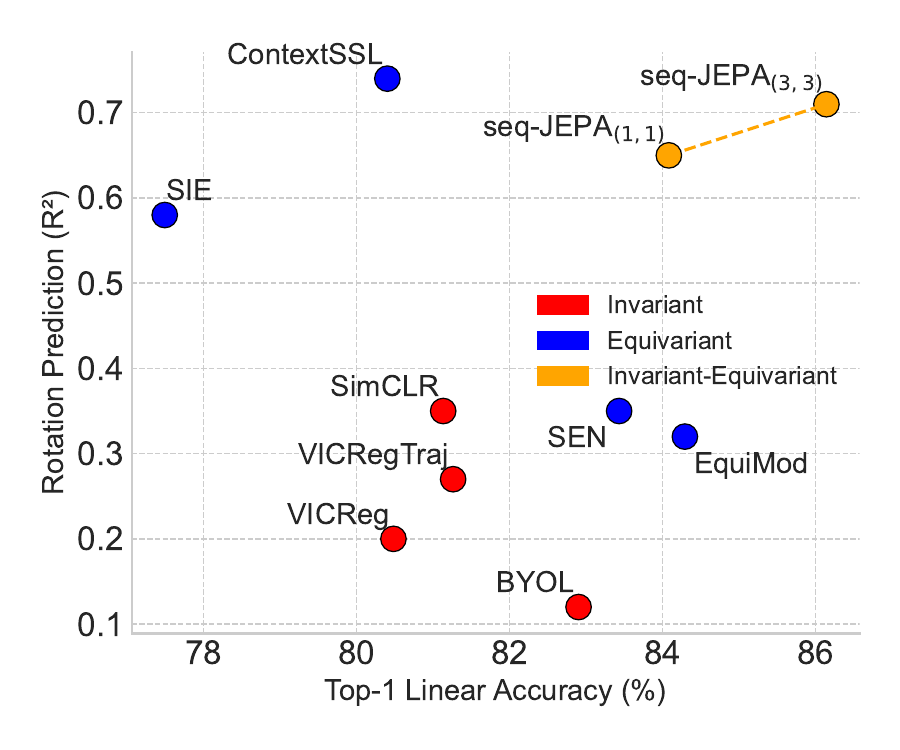}
    \caption{\small Top-1 linear classification (invariance) vs. rotation prediction (equivariance) performance on 3DIEBench. seq-JEPA learns good representations for both tasks. Subscripts indicate training and inference sequence lengths.}
    \label{fig:inv-equi-plot}
  \end{minipage}
  \hfill
  \begin{minipage}{0.55\linewidth}
    \centering
    \captionof{table}{\small Evaluation on 3DIEBench for linear probe classification (invariance-demanding) and rotation prediction (equivariance-demanding). Equivariant models and seq-JEPA are conditioned on rotation. For seq-JEPA, training and inference sequence lengths are denoted  by $M_{tr}$ and $M_{val}$.}
  \resizebox{0.98\linewidth}{!}{
    \begin{tabular}{lccc}
        \toprule
        Method & Top-1 Acc. (\%) & Rel. Rot. ($R^2$) & Indiv. Rot. ($R^2$) \\
        \midrule
        \textit{\textcolor{red}{Invariant}} & & & \\
        BYOL & 82.90 & 0.12 & 0.25 \\
        SimCLR & 81.13  &  0.35 & 0.54 \\
        VICReg &  80.48 & 0.20 & 0.36 \\
        VICRegTraj  & 81.26	& 0.27  &	0.43 \\
        \midrule
        \textit{\textcolor{blue}{Equivariant}} & & & \\
        SEN  & 83.43  &   0.35 & 0.57 \\
        EquiMod &  84.29 &  0.32 & 0.55 \\
        SIE  & 77.49 & 0.58 & 0.62 \\
        Conditional BYOL  & 82.61 & 0.31 & 0.47 \\
        ContextSSL ($c=126$) & 80.40 &  \textbf{0.74} & \textbf{0.78} \\
        \midrule
        \textit{\textcolor{orange}{Invariant-Equivariant}} & & & \\
        \rowcolor{Light} seq-JEPA (1, 1) & 84.08 & 0.65 & 0.69 \\
        \rowcolor{Light} seq-JEPA (1, 3) & \underline{85.31} & 0.65 & 0.69 \\
        \rowcolor{Light} seq-JEPA (3, 3) & 86.14 & \underline{0.71} & \underline{0.74} \\
        \rowcolor{Light} seq-JEPA (3, 5) & \textbf{87.41} & \underline{0.71} & \underline{0.74} \\
        \rowcolor{Light} seq-JEPA (no act cond) & 86.05 & 0.29 & 0.37 \\
        \bottomrule
    \end{tabular}
    }
    \label{tab:3db-rot}
  \end{minipage}
\end{figure*}

We use the 3DIEBench benchmark to quantitatively compare performance on equivariance- and invariance-demanding downstream tasks in seq-JEPA with baseline methods. This benchmark allows us to measure equivariance through decoding 3D object rotations while enabling invariance measurement through object classification. Table~\ref{tab:3db-rot} provides a summary of our evaluation on the 3DIEBench where equivariant methods have been conditioned on rotation (a 4-D quaternion representing the relative rotation between two views). In addition to the relative rotation between two views, in the last column we provide the $R^2$ score for predicting individual transformation parameters from representations of a single view. For seq-JEPA, we trained models with varying training sequence lengths (denoted by $M_{tr}$ in the table) and measure the linear classification performance on top of aggregate representations with different inference lengths (denoted by $M_{val}$).

Among invariant methods, BYOL achieves the highest classification accuracy, yet does not offer a high level of equivariance. Adding a linear trajectory regularization loss to VICReg~\citet{wang2024pose} without using ground-truth transformations improves over VICReg, which shows that imposing geometric priors can improve both invariant and equivariant performance---even when these priors do not fully materialize in the environment, e.g., when we have non-smooth angle changes as in 3DIEBench. Among equivariant baselines, SIE and ContextSSL yield strong rotation prediction performance due to their specialized equivariance predictors and loss functions, but underperform in classification. EquiMod and SEN offer better classification performance, yet compromise equivariance. In contrast, seq-JEPA achieves strong performance in both, matching the best rotation $R^2$ scores while exceeding baselines in classification, with gains increasing with inference sequence length $M_{val}$. Ablating action conditioning leads to a sharp drop in equivariance but retains classification accuracy, confirming our hypothesis that action-conditioned sequential aggregation enables a distinct representational structure for invariance- and equivariance-demanind tasks. For additional results on 3DIEBench including MRR, Hit@1, and Hit@5 metrics, performance of models with varying training and inference sequence lengths and models conditioned on both rotation and color, and evaluation on an out-of-distribution (OOD) set of 3DIEBench see Appendices~\ref{sec:rotcol} to~\ref{sec:heatmaps}.

\subsection{Qualitative evaluation on 3DIEBench}

\begin{figure*}[t]
  \centering
  \begin{minipage}{0.49\linewidth}
    \centering
    \includegraphics[width=\linewidth]{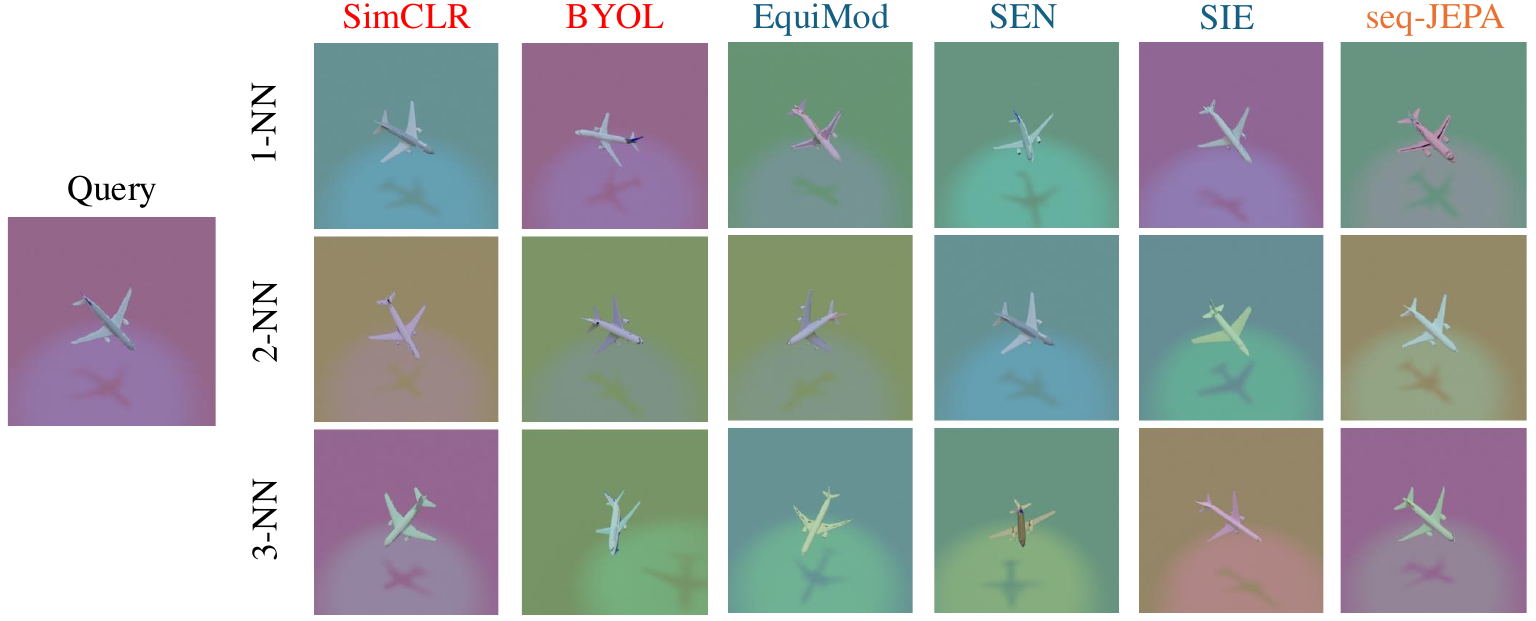}
    \caption{\small Retrieval of nearest representations; given a query image, we extract the three nearest encoder representations in the validation set of 3DIEBench. The retrieved views of models with the highest quantitative rotation equivariance performance maintain the rotation of the query image across all retrieved views.}
    \label{fig:nn-3db}
  \end{minipage}
  \hfill
  \begin{minipage}{0.49\linewidth}
    \centering
    \includegraphics[width=\linewidth]{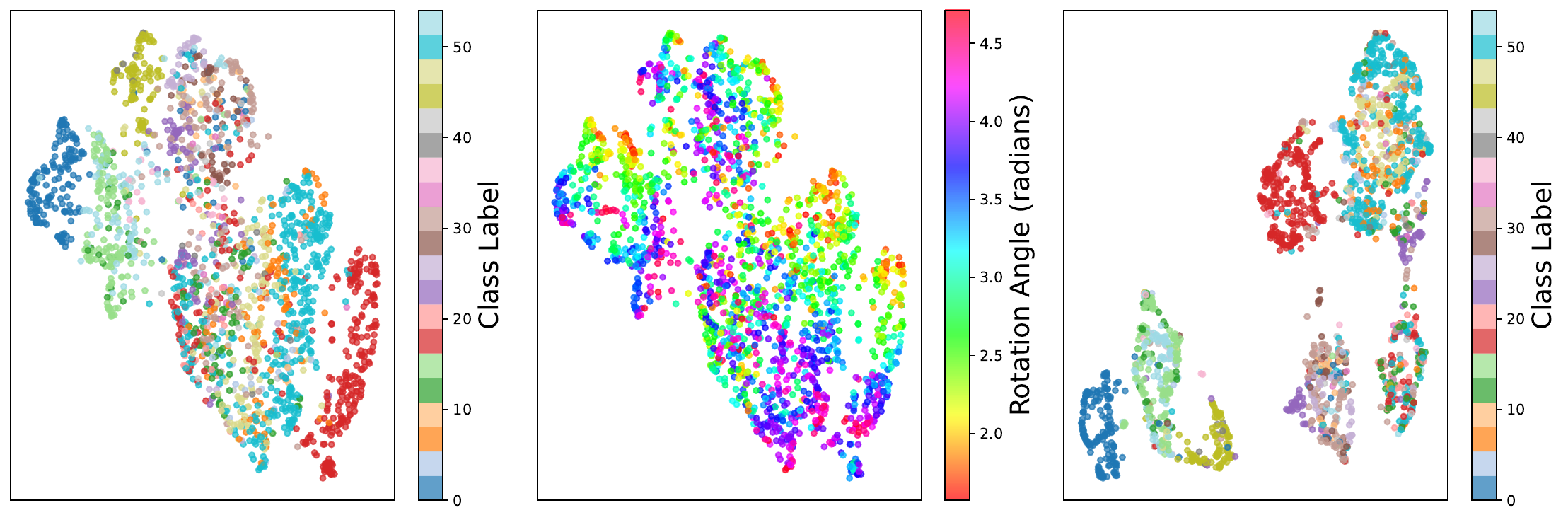}
    \caption{\small 2-D UMAP projections of seq-JEPA's encoder and aggregate representations trained on 3DIEBench and conditioned on rotation with $M_{tr}=3$ and $M_{val}=5$. Encoder representations for each view observation, color-coded by class (\textbf{left}) and rotation angle (\textbf{middle}). Aggregate representation for \( M_{val} = 5 \), color-coded by class (\textbf{right}).}
    \label{fig:umap-3db}
  \end{minipage}
\end{figure*}

To visualize equivariance in representational space, we retrieve the three nearest representations of a query image from the validation set of 3DIEBench (Figure~\ref{fig:nn-3db}). While all models retrieve the correct object category, only seq-JEPA and SIE consistently preserve rotation across all retrieved views, consistent with their high $R^2$ scores. Next, we projected encoder and aggregate representations using 2D UMAP (Figure~\ref{fig:umap-3db}). The left panel shows encoder representations colored by class label, while the middle panel displays the same encoder representations colored by rotation angle. The smooth color gradation across the map within each class cluster in the middle panel suggests that the encoder captures rotation angle as a continuous factor, implying equivariance to rotation (e.g., the red class in the bottom-right corner of the right panel and the corresponding part in the middle panel). The right panel shows aggregate representations colored by class label. Comparing the class-colored plots (left and right panels), we observe that both encoder and aggregate representations contain class information. However, when we aggregate multiple views of a sample, some of the intra-class variability (resulting from transformations such as rotation) is eliminated, causing each class' representational cluster to become more homogeneous. This aggregation procedure likely reduces variation due to rotation and makes the representations more invariant, resulting in decreased intra-class spread and increased inter-class distance. We create a similar UMAP visualization for seq-JEPA with ablated rotation conditioning in Appendix~\ref{sec:umap} to highlight the role of action conditioning in achieving equivariance to rotation.

\subsection{Evaluation with Hand-Crafted Augmentations}

We assess downstream performance under hand-crafted augmentations by training on CIFAR100 and Tiny ImageNet (Table~\ref{tab:cifar-tinyimg}). Models with action conditioning are trained conditioned on crop, color jitter, blur, or all three (indicated in the first column of the table). seq-JEPA consistently achieves higher equivariance than both invariant and equivariant baselines across all transformations. Notably, except for the model trained on CIFAR-100 and conditioned on blur, the best equivariance performance for a given augmentation is achieved when the model is specialized and conditioned only on that augmentation. Furthermore, ablating actions (last row in the table) causes seq-JEPA to lose its equivariance across transformations compared to action-conditioned models. Overall, our model outperforms both invariant and equivariant families in terms of equivariance, while being competitive in terms of classification performance. For additional results with varying training and inference sequence lengths, see Appendix~\ref{sec:heatmaps}.

\begin{table*}[!t]
    \caption{\small Evaluation with hand-crafted augmentations on CIFAR100 and Tiny ImageNet; equivariance is measured by predicting relative transformation parameters associated with crop, color jitter, or blur augmentations. For all seq-JEPA models, $M_{val}=5$.}
    \label{tab:cifar-tinyimg}
    \centering
\resizebox{0.93\linewidth}{!}{
\centering
    \begin{tabular}{c|lcccc||cccc}
    \toprule
    & & \multicolumn{4}{c}{\textbf{CIFAR100}} & \multicolumn{4}{c}{\textbf{Tiny ImageNet}} \\
    Conditioning & Method & Classification (top-1) & Crop ($R^2$) & Jitter ($R^2$) & Blur ($R^2$)  
      & Classification (top-1) & Crop ($R^2$) & Jitter ($R^2$) & Blur ($R^2$) \\
    \midrule
    \midrule
    &  \textit{\textcolor{red}{Invariant}}&   &   &  &  &  &  &  &  \\
   \multirow{3}{*}{-} 
   & SimCLR & 61.72 &  0.56 & 0.18 & 0.04 & 34.29 & 0.30 & 0.08 & 0.13 \\
   & BYOL & 62.17 & 0.39 & 0.05 & -0.01 & 35.70 & 0.17 & 0.01 & 0.16 \\
   & VICReg & 61.35 & 0.49 & 0.11 & -0.06 & 35.29 & 0.31 & 0.05 & 0.19 \\
   & VICRegTraj & 61.07 & 0.46 & 0.14 & 0.02 & 34.95 & 0.28 & 0.09 & 0.16 \\
    \thickmidrule
    & \textit{\textcolor{blue}{Equivariant}}&   &   &  &  &  &  &  &  \\
   \multirow{4}{*}{\tiny{Crop+Jitter+Blur}}  
   & SEN  & 61.94 & 0.65 & 0.51 & 0.85  & 36.01 & 0.24 & 0.48 & 0.87 \\
   & EquiMod & 61.80  & 0.59  & 0.49 & 0.74 & 36.75 & 0.38 & 0.46 & 0.86 \\
   & SIE  & 58.81 &  0.34  &  0.26 & 0.53 & 31.37 & 0.26 & 0.56 & 0.88 \\
   & Conditional BYOL  & 60.63 & 0.56 & 0.46 & 0.73 & 35.49 & 0.34 & 0.54 & 0.87 \\
   \thickmidrule
   \multirow{4}{*}{Crop}  
   & SEN  & 61.56 & 0.66 & 0.15 & 0.10  & 35.95 & 0.24 & 0.11 & 0.49 \\
   & EquiMod & 61.77  & 0.62  & 0.15 & 0.01 & 36.83 & 0.26 & 0.12 & 0.30 \\
   & SIE  & 57.55 & 0.69  & 0.11 & 0.25 & 32.38 & 0.34 & 0.05 & 0.14 \\
   & Conditional BYOL  & 60.17 & 0.55  & 0.10 & -0.01 & 37.07 & 0.22 & 0.01 & 0.24 \\
   \thickmidrule
   \multirow{4}{*}{Color jitter} 
   & SEN  & 61.78 & 0.50 & 0.52 & 0.02  & 36.59 & 0.26 & 0.50 & 0.21 \\
   & EquiMod & 61.53  & 0.44  & 0.50 & -0.02 & 37.18 & 0.20 & 0.52 & 0.29 \\
   & SIE  & 59.29 & 0.48  & 0.59 & 0.06 & 34.37 & 0.39 & 0.62 & 0.35 \\
   & Conditional BYOL  & 61.30 & 0.36 & 0.52 & 0.02  & \textbf{37.93} & 0.24 & 0.62 & 0.30 \\
   \thickmidrule
   \multirow{4}{*}{Blur} 
   & SEN  & 61.47 & 0.43 & 0.15 & 0.84  & 34.62 & 0.16 & 0.08 & 0.79 \\
   & EquiMod & \textbf{62.72}  & 0.41  & 0.15 & 0.74 & 36.12 & 0.24 & 0.10 & 0.91 \\
   & SIE  & 57.66 & 0.40  & 0.07 & 0.71 & 31.00 & 0.26 & 0.05 & 0.85 \\
   & Conditional BYOL  & 60.74 & 0.36 & 0.11 & 0.69  & 36.17 & 0.19 & 0.02 & 0.85 \\
    \thickmidrule
    & \textit{\textcolor{orange}{Invariant-Equivariant}}&  &   &  &  &  &  &  &  \\
\rowcolor{Light}
 \tiny{Crop+Jitter+Blur}   & seq-JEPA ($M_{tr}=1$) & 52.90 & 0.77  &  0.52 & 0.23 & 37.10 & 0.64 & 0.49 & 0.89 \\
     \rowcolor{Light}
 \tiny{Crop+Jitter+Blur}   & seq-JEPA ($M_{tr}=2$) & 60.17 & 0.78  &  0.64 & 0.88 & 35.56 & 0.69 & 0.42 & 0.93 \\
       \rowcolor{Light}
 \tiny{Crop+Jitter+Blur}  & seq-JEPA ($M_{tr}=3$) & 58.33 & \textbf{0.79}  & 0.63 & \textbf{0.92} & 34.85 & 0.67 & 0.64 & 0.96 \\
      \rowcolor{Light}
 Crop & seq-JEPA ($M_{tr}=2$) & 59.32 & 0.78  & 0.01 & 0.10 & 35.74 & \textbf{0.70} & 0.12 & 0.46 \\
 \rowcolor{Light}
  Color Jitter & seq-JEPA ($M_{tr}=3$) & 58.62 & 0.68 & \textbf{0.68} & 0.29 & 35.21 & 0.60 & \textbf{0.66} & 0.62 \\
 \rowcolor{Light}
   Blur   & seq-JEPA ($M_{tr}=3$) & 56.82 & 0.71 & 0.15 & 0.74  & 35.79 & 0.58 & 0.22 & \textbf{0.97} \\
\rowcolor{Light}
 - & seq-JEPA ($M_{tr}=2$) & 58.37 & 0.64 & 0.14 & 0.16 & 35.97 & 0.52 & 0.18 & 0.47 \\
     \bottomrule
    \end{tabular}
    }
\end{table*}


\subsection{Predictive Learning across Saccades and Path Integration}
\begin{figure*}[t]
  \centering
  \begin{minipage}{0.51\linewidth}
    \centering
    \captionof{table}{\small Evaluation of predictive learning across saccades on STL-10. Equivariance is measured with respect to fixation coordinates. Unless stated otherwise, $M_{tr} = M_{val} = 4$.}

    \resizebox{0.92\linewidth}{!}{
    \begin{tabular}{c|lcc}
    \toprule
    Conditioning &  & Top-1 Acc. (\%) & Position ($R^2$)  \\
    \midrule
    & \textit{\textcolor{red}{Invariant}} & & \\
    - & SimCLR (augmentations) & 85.23 & -0.06 \\
    \thickmidrule
    & \textit{\textcolor{blue}{Equivariant}} & & \\
    position & Conv-JEPA ($M_{tr,val}=4$) & 80.04 & 0.80 \\
    \thickmidrule
    & \textit{\textcolor{orange}{Invariant-Equivariant}} & & \\
    \rowcolor{Light} - & seq-JEPA & 70.45 & 0.38 \\
    \rowcolor{Light} position & seq-JEPA & 83.44 & 0.80 \\
    \rowcolor{Light} position & seq-JEPA ($M_{val}=6$) & 84.12 & 0.80 \\
    \rowcolor{Light} position & seq-JEPA (w/o saliency \& IoR) & 79.85 & 0.88 \\
    \rowcolor{Light} position & seq-JEPA (w/o IoR) & 77.97 & 0.85 \\
    \bottomrule
    \end{tabular}
    }
    \label{tab:stl-position}
  \end{minipage}
  \hfill
  \begin{minipage}{0.46\linewidth}
    \centering
    \includegraphics[width=\linewidth]{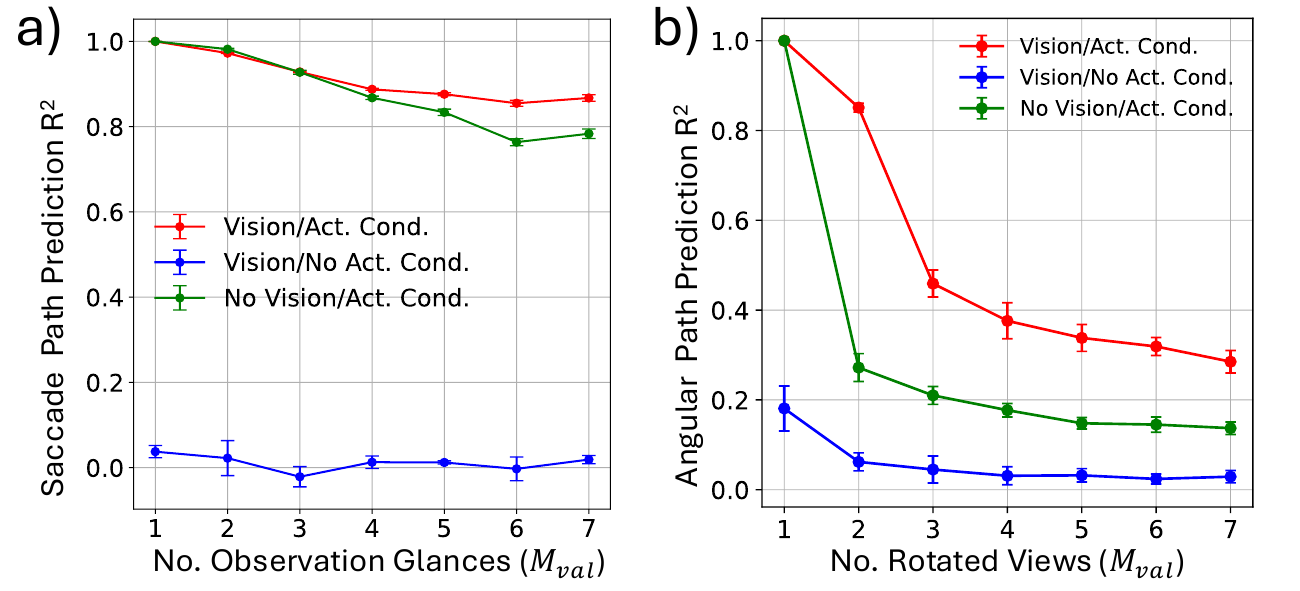}
    \caption{\small a) Visual path integration across eye movements, b) Angular path integration across object rotations (results over three random seeds)}
    \label{fig:path-int}
  \end{minipage}
\end{figure*}

In our third action-observation setting, we consider predictive learning across simulated eye movements to exhibit seq-JEPA's ability in leveraging a sequence of partial observations to learn visual representations. In Table~\ref{tab:stl-position}, seq-JEPA reaches 83.44\% top-1 accuracy on STL-10, comparable to SimCLR (85.23\%) trained with full-resolution images and strong augmentations. This gap narrows further when increasing the inference sequence length from $M_{val}=4$ to $M_{val}=6$. Ablating action conditioning causes the accuracy on top of the aggregate representations to drop sharply, indicating that 2-D positional awareness is essential to forming semantic representations across simulated eye movements. Compared to Conv-JEPA---which accumulates prediction losses pairwise---seq-JEPA performs better in classification, highlighting the importance of sequence aggregation when dealing with partial observations in SSL. Further ablations show that saliency-driven sampling and IoR are critical for forming informative, non-overlapping patch sequences and subsequently a high-quality aggregate representation. Interestingly, while random uniform patch sampling negatively impacts classification accuracy due to lower semantic content, it results in the highest positional equivariance as the model samples patches and corresponding saccade actions from a more diverse set of positions across the entire image, not just the salient regions. The UMAP projections for PLS with and without action conditioning in Appendix~\ref{sec:umap} further underscore the role of action conditioning in enabling positional equivariance.

\textbf{Path Integration.} In the context of eye movement-driven or any sequential observations, an ability that naturally arises from predictive learning is \textit{path integration}~\citep{mcnaughton_path_2006}, i.e., predicting the cumulative transformation/action from a sequence of actions. We evaluate this task in both eye movements in PLS (visual path integration) and object rotations in 3DIEBench (angular path integration). As shown in Figure~\ref{fig:path-int}, seq-JEPA demonstrates strong performance in both settings, with performance degrading gracefully as sequence length increases. Ablating action conditioning causes path integration to fail, whereas ablating the visual stream has only a minor impact—highlighting that action information is the dominant signal for this task. Full details of the path integration setup are provided in Appendix~\ref{sec:pathint}.

\subsection{Action Conditioning Ablations}
\label{sec:ablation}

\begin{figure*}[t]
  \centering
  \begin{minipage}{0.42\linewidth}
    \centering
    \captionof{table}{\small Ablation results for action conditioning (3DIEBench). All models use $M_{tr}=3$, $M_{val}=5$ (results over three random seeds).}

    \resizebox{0.9\linewidth}{!}{
    \begin{tabular}{lcc}
        \toprule
        Variant & Top-1 Acc. (\%) & Rotation ($R^2$) \\
        \midrule
        \textbf{Act. conditioning} & & \\
        None & 87.36 $\pm$ 0.7 & 0.29 $\pm$ 0.04 \\
        No predictor cond. & 87.17 $\pm$ 0.3 & 0.37 $\pm$ 0.06 \\
        No transformer cond. & 86.33 $\pm$ 0.1 & 0.53 $\pm$ 0.05 \\
        \midrule
        \textbf{Act. embedding dim.} & & \\
        $a_{\text{dim}}=16$ & 86.29 $\pm$ 0.4 & 0.70 $\pm$ 0.01 \\
        $a_{\text{dim}}=64$ & 87.11 $\pm$ 0.2 & 0.70 $\pm$ 0.02 \\
        $a_{\text{dim}}=128$ (default) & \textbf{87.41} $\pm$ 0.5 & 0.71 $\pm$ 0.02 \\
        $a_{\text{dim}}=256$ & 87.26 $\pm$ 0.6 & \textbf{0.72} $\pm$ 0.00 \\
        \bottomrule
    \end{tabular}
    }
    \label{tab:3diebench-ablation}
  \end{minipage}
  \hfill
  \begin{minipage}{0.55\linewidth}
    \centering
    \includegraphics[width=\linewidth]{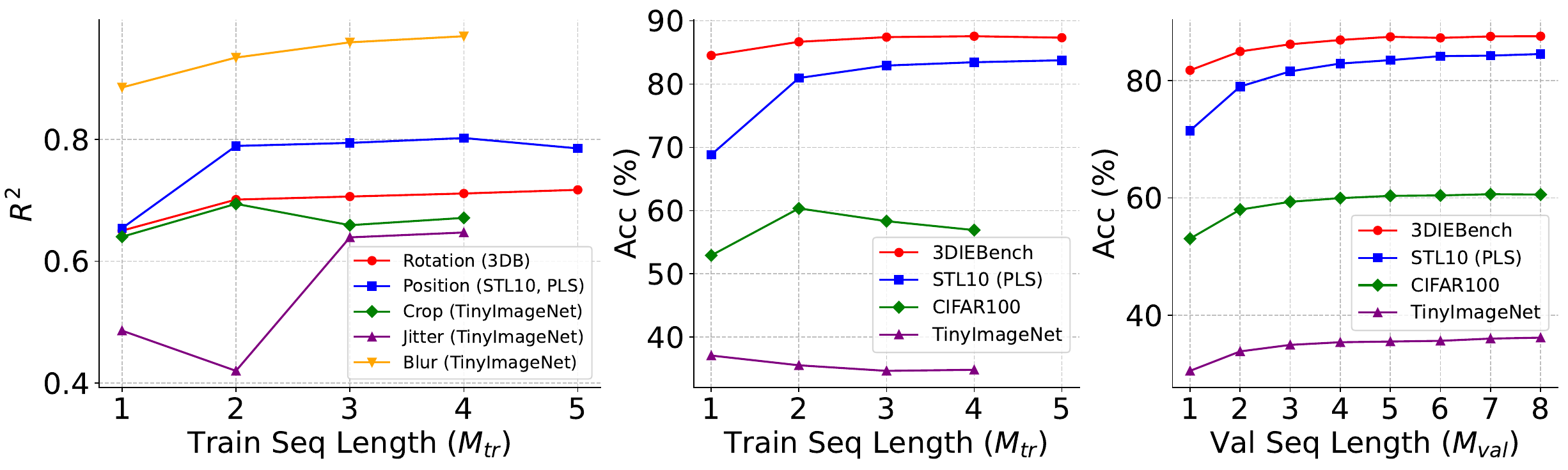}
    \caption{\small Effect of training and inference sequence length on seq-JEPA's performance; \textbf{left:}: Equivariant performance ($R^2$) versus training sequence length; \textbf{middle:} Classification accuracy versus training sequence length; \textbf{right:} Classification accuracy versus inference sequence length.}
    \label{fig:seq-len}
  \end{minipage}
\end{figure*}

To better understand the mechanisms underlying seq-JEPA's invariant-equivariant representation learning and role of action conditioning, we perform a set of ablation experiments on 3DIEBench. Specifically, we study: (i) the role of action conditioning in the transformer and predictor; (ii) the impact of action embedding dimensionality. Table~\ref{tab:3diebench-ablation} summarizes our ablation results. Removing action conditioning entirely causes a significant drop in equivariance ($R^2$ from 0.71 to 0.29), although classification accuracy remains high thanks to sequence aggregation and segregated invariance-equivariance in our model. Conditioning only the transformer or only the predictor leads to intermediate results, with predictor conditioning proving more critical for equivariance. We also vary the dimensionality of the learnable action embeddings: performance saturates around the default size of 128, with smaller sizes (e.g., 16 or 64) already sufficient to capture the rotation structure.

\subsection{Scaling Properties: Role of Training and Inference Sequence Lengths} \label{sec:seqlen}

We study the effect of both training and inference sequence lengths on performance across tasks, i.e., scalability of seq-JEPA in terms of context length (Figure~\ref{fig:seq-len}). We draw inspiration from recent findings in foundation models, where increased training and inference context—whether in text~\citep{brown_language_2020,touvron_llama_2023}, vision~\citep{zellers_merlot_2021,chen_empirical_2021}, or video~\citep{bain_frozen_2021,arnab_vivit_2021}—consistently leads to stronger representations.

We observe that equivariance generally improves with longer training sequences (left panel). A possible explanation for this observation is the presence of more transitions $(z_i, a_i, z_{i+1})$ in working memory, which means the predictor has access to a richer context $z_{\mathrm{agg}}$. This enables the predictor to more accurately approximate the transition $p(z_{i+1} \mid z_{\mathrm{agg,i}}, a_i)$. Because accurate prediction requires the model to preserve and utilize information about $a_t$, the encoder is implicitly encouraged to learn structured, more equivariant representations.

Classification performance on 3DIEBench and STL-10 (middle panel) benefits from a longer training sequence. In contrast, on CIFAR100 and Tiny ImageNet with synthetic augmentations, longer training sequences slightly decrease classification performance. We hypothesize that leveraging and aggregating a sequence of action-observation pairs, i.e. seq-JEPA's architectural inductive bias, is most effective in settings where the downstream task benefits from sequential observations. In the case of object rotations in 3DIEBench, seeing an object from multiple angles is indeed beneficial in recognizing the object's category, which explains the improved classification accuracy with increased training sequence length. Similarly, in the case of predictive learning across saccades, each eye movement and its subsequent glance provides additional information that can be leveraged for learning a richer aggregate representation. 

At inference time, all datasets benefit from longer context lengths ($M_{val}$), confirming that richer aggregate representations yield stronger performance (right plot). This scalability via sequence lengths opens avenues for efficient representation learning with small foveated patches in lieu of full-frame inputs, mirroring how foundation models scale with input tokens at test time. Together with our transfer learning results on ImageNet-1k (Appendix~\ref{sec:transfer}), these findings suggest that seq-JEPA's architectural inductive bias enables graceful scaling via longer sequence lengths.

\section{Related Work}
\textbf{Non-Generative World Models and Joint-Embedding Predictive Architectures.}  
Non-generative world models predict the effect of transformations or actions directly in latent space, avoiding reconstruction in pixel space. This includes contrastive SSL methods that model transformed views from context representations~\citep{van_den_oord_representation_2019,gupta_-context_2024}, as well as approaches in model-based reinforcement learning (RL) to improve sample efficiency, generate intrinsic rewards, or capture environment transitions~\citep{schwarzer_data-efficient_2021,khetarpal_unifying_2025,ni_bridging_2024,tang_understanding_2023,guo_byol-explore_2022}. Joint-embedding predictive architectures~\citep{lecun_path_2022} form a subclass of non-generative world models. They introduce an asymmetric predictor conditioned on transformation parameters to infer the outcome of an action applied to a latent view. Examples include I-JEPA\citep{assran_self-supervised_2023}, which predicts masked regions from positional cues, and IWM~\citep{garrido_learning_2024}, which conditions on augmentation parameters. JEPAs have been recently extended to physical reasoning~\citep{garrido_intuitive_2025} and offline planning~\citep{sobal_learning_2025}, illustrating the framework’s versatility in representation learning and world modeling.

\textbf{Equivariant SSL.}  
Equivariant SSL methods aim to retain transformation-specific information in the latent space, typically by augmenting invariant objectives with an additional equivariance term. Some approaches directly predict transformation parameters~\citep{lee_improving_2021,scherr_self-supervised_2022,gidaris_unsupervised_2018,gupta_-context_2024,dangovski_equivariant_2022}. Methods such as EquiMod~\citep{devillers_equimod_2022} and SIE~\citep{garrido_self-supervised_2023} predict the effect of a transformation in latent space via a predictor in addition to their invariant objective. SEN~\citep{park_learning_2022} similarly predicts transformed representations but omits the invariance term. \citet{xiao_what_2021} use contrastive learning with separate projection heads for each augmentation, treating same-augmentation pairs as negatives. ContextSSL~\citep{gupta_-context_2024} conditions representations on both current actions and recent context and employs a dual predictor for transformation prediction to avoid collapsing to invariance. Other approaches~\citep{shakerinava_structuring_2022,gupta_structuring_2023,yerxa_contrastive-equivariant_2024} do not require explicit transformation parameters but instead enforce equivariance by applying the same transformation to multiple view pairs and minimizing a distance-based loss. Action-conditioned JEPAs incorporate augmentation parameters and mask positions by conditioning the predictor to induce equivariance without additional objectives~\citep{garrido_learning_2024}. \citet{chavhan2023quality} use an ensemble of heads trained on top of a pre-trained SSL encoder to span a diverse spectrum of transformation sensitivities across the latent spaces of each head. Downstream probes then learn to linearly combine these feature heads depending on the desired invariance-equivariance trade-off.

\textbf{Positioning of Our Work.}  
seq-JEPA belongs to the family of joint-embedding predictive architectures and is a non-generative world model. Unlike most equivariant SSL methods, seq-JEPA does not rely on an equivariance loss or transformation prediction objective, nor does it require view pairs with matched transformations. Instead, it leverages action conditioning and architectural inductive biases to learn two separate invariant and equivariant representations. In contrast to ContextSSL, which extends the two-view contrastive setting of contrastive predictive coding~\citet{van_den_oord_representation_2019} using a transformer decoder projector conditioned on previous views via key-value caching, seq-JEPA operates on sequences of action-observation pairs in an online end-to-end manner by incorporating a sequence model (e.g. a transformer encoder) as a \emph{learned working memory} during both training and inference. Moreover, while ContextSSL aims to adapt equivariance to recent transformations by dynamically modifying the training distribution, seq-JEPA is designed to explicitly learn both invariant and equivariant representations with respect to a specified set of transformations, and is also well-suited for downstream tasks that require multi-step observation aggregation.

\section{Limitations and Future Perspectives} \label{sec:lim}

We have validated viability of seq-JEPA across a range of transformations in the image domain. Here, we discuss limitations and possible future directions. First, seq-JEPA is capable of autoregressive prediction in time, and therefore, can be leveraged for autoregressive latent planning in control tasks. Second, the transformer-based aggregator in seq-JEPA could support multi-modal fusion across language, audio, or proprioceptive inputs—enabling multi-modal world modeling and generalization. Third, while our results show clear benefits from longer sequences, we have experimented with relatively short training and inference sequence lengths as the backbone is also trained end-to-end. The sequence scaling trends observed in the paper suggest that seq-JEPA can benefit from longer context windows over representations of pre-trained backbones as in~\citep{pang_frozen_2023,lin_vedit_2024}. Fourth, our method assumes access to a known transformation group (e.g., $SO(3)$ for 3D rotations). Designing group-agnostic or learned transformation models~\citep{finzi_practical_2021} without access to transformation parameters or pairs of same transformation is an open challenge in equivariant SSL that future work may tackle. Finally, our preliminary ImageNet-1k transfer results (Appendix~\ref{sec:transfer}) point to potential for broader generalization. Scaling seq-JEPA to larger foveated image settings or video and multi-modal datasets such as Ego4D \citet{song_ego4d_2023} could support the development of lightweight, saliency-driven agents capable of learning efficiently from partial observations in embodied settings with a limited field of vision.

\begin{ack}
This project was supported by funding from NSERC (Discovery Grants RGPIN-2022-05033 to E.B.M., and RGPIN-2023-03875 to S.B.), Canada CIFAR AI Chairs Program and Google to E.B.M., Canada Excellence Research Chairs (CERC) Program, Mila - Quebec AI Institute, Institute for Data Valorization (IVADO), CHU Sainte-Justine Research Centre, Fonds de Recherche du Québec–Santé (FRQS), and a Canada Foundation for Innovation John R. Evans Leaders Fund grant to E.B.M. This research was also supported in part by Digital Research Alliance of Canada (DRAC) and Calcul Québec.
\end{ack}



\bibliography{references}

@inproceedings{pang_frozen_2023,
	title = {Frozen {Transformers} in {Language} {Models} {Are} {Effective} {Visual} {Encoder} {Layers}},
	url = {https://openreview.net/forum?id=t0FI3Q66K5},
	language = {en},
	urldate = {2025-05-15},
	author = {Pang, Ziqi and Xie, Ziyang and Man, Yunze and Wang, Yu-Xiong},
	month = oct,
	year = {2023},
}

@inproceedings{chavhan2023quality,
  title={Quality Diversity for Visual Pre-Training},
  author={Chavhan, Ruchika and Gouk, Henry and Li, Da and Hospedales, Timothy},
  booktitle={Proceedings of the IEEE/CVF International Conference on Computer Vision},
  pages={5384--5394},
  year={2023}
}

@misc{rusak_infonce_2025,
	title = {{InfoNCE}: {Identifying} the {Gap} {Between} {Theory} and {Practice}},
	shorttitle = {{InfoNCE}},
	url = {http://arxiv.org/abs/2407.00143},
	doi = {10.48550/arXiv.2407.00143},
	urldate = {2025-05-15},
	publisher = {arXiv},
	author = {Rusak, Evgenia and Reizinger, Patrik and Juhos, Attila and Bringmann, Oliver and Zimmermann, Roland S. and Brendel, Wieland},
	month = apr,
	year = {2025},
	note = {arXiv:2407.00143 [cs]},
}

@inproceedings{finzi_practical_2021,
	title = {A {Practical} {Method} for {Constructing} {Equivariant} {Multilayer} {Perceptrons} for {Arbitrary} {Matrix} {Groups}},
	url = {https://proceedings.mlr.press/v139/finzi21a.html},
	language = {en},
	urldate = {2025-05-15},
	booktitle = {Proceedings of the 38th {International} {Conference} on {Machine} {Learning}},
	publisher = {PMLR},
	author = {Finzi, Marc and Welling, Max and Wilson, Andrew Gordon},
	month = jul,
	year = {2021},
	note = {ISSN: 2640-3498},
	pages = {3318--3328},
}

@inproceedings{bain_frozen_2021,
	address = {Montreal, QC, Canada},
	title = {Frozen in {Time}: {A} {Joint} {Video} and {Image} {Encoder} for {End}-to-{End} {Retrieval}},
	copyright = {https://doi.org/10.15223/policy-029},
	isbn = {978-1-6654-2812-5},
	shorttitle = {Frozen in {Time}},
	url = {https://ieeexplore.ieee.org/document/9711165/},
	doi = {10.1109/ICCV48922.2021.00175},
	language = {en},
	urldate = {2025-05-15},
	booktitle = {2021 {IEEE}/{CVF} {International} {Conference} on {Computer} {Vision} ({ICCV})},
	publisher = {IEEE},
	author = {Bain, Max and Nagrani, Arsha and Varol, Gul and Zisserman, Andrew},
	month = oct,
	year = {2021},
	pages = {1708--1718},
}

@inproceedings{arnab_vivit_2021,
	address = {Montreal, QC, Canada},
	title = {{ViViT}: {A} {Video} {Vision} {Transformer}},
	copyright = {https://doi.org/10.15223/policy-029},
	isbn = {978-1-6654-2812-5},
	shorttitle = {{ViViT}},
	url = {https://ieeexplore.ieee.org/document/9710415/},
	doi = {10.1109/ICCV48922.2021.00676},
	language = {en},
	urldate = {2025-05-15},
	booktitle = {2021 {IEEE}/{CVF} {International} {Conference} on {Computer} {Vision} ({ICCV})},
	publisher = {IEEE},
	author = {Arnab, Anurag and Dehghani, Mostafa and Heigold, Georg and Sun, Chen and Lucic, Mario and Schmid, Cordelia},
	month = oct,
	year = {2021},
	pages = {6816--6826},
}

@misc{zellers_merlot_2021,
	title = {{MERLOT}: {Multimodal} {Neural} {Script} {Knowledge} {Models}},
	shorttitle = {{MERLOT}},
	url = {http://arxiv.org/abs/2106.02636},
	doi = {10.48550/arXiv.2106.02636},
	urldate = {2025-05-15},
	publisher = {arXiv},
	author = {Zellers, Rowan and Lu, Ximing and Hessel, Jack and Yu, Youngjae and Park, Jae Sung and Cao, Jize and Farhadi, Ali and Choi, Yejin},
	month = oct,
	year = {2021},
	note = {arXiv:2106.02636 [cs]},
}

@misc{touvron_llama_2023,
	title = {{LLaMA}: {Open} and {Efficient} {Foundation} {Language} {Models}},
	shorttitle = {{LLaMA}},
	url = {http://arxiv.org/abs/2302.13971},
	doi = {10.48550/arXiv.2302.13971},
	urldate = {2025-05-15},
	publisher = {arXiv},
	author = {Touvron, Hugo and Lavril, Thibaut and Izacard, Gautier and Martinet, Xavier and Lachaux, Marie-Anne and Lacroix, Timothée and Rozière, Baptiste and Goyal, Naman and Hambro, Eric and Azhar, Faisal and Rodriguez, Aurelien and Joulin, Armand and Grave, Edouard and Lample, Guillaume},
	month = feb,
	year = {2023},
	note = {arXiv:2302.13971 [cs]},
}

@misc{garrido_intuitive_2025,
	title = {Intuitive physics understanding emerges from self-supervised pretraining on natural videos},
	url = {http://arxiv.org/abs/2502.11831},
	doi = {10.48550/arXiv.2502.11831},
	urldate = {2025-03-06},
	publisher = {arXiv},
	author = {Garrido, Quentin and Ballas, Nicolas and Assran, Mahmoud and Bardes, Adrien and Najman, Laurent and Rabbat, Michael and Dupoux, Emmanuel and LeCun, Yann},
	month = feb,
	year = {2025},
	note = {arXiv:2502.11831 [cs]},
}

@inproceedings{assran_masked_2022,
	title = {Masked {Siamese} {Networks} for {Label}-{Efficient} {Learning}},
	url = {http://arxiv.org/abs/2204.07141},
	language = {en},
	urldate = {2024-05-21},
	booktitle = {Proceedings of the 17th {European} {Conference} on {Computer} {Vision}, {ECCV} 2022},
	author = {Assran, Mahmoud and Caron, Mathilde and Misra, Ishan and Bojanowski, Piotr and Bordes, Florian and Vincent, Pascal and Joulin, Armand and Rabbat, Michael and Ballas, Nicolas},
	month = apr,
	year = {2022},
	note = {arXiv:2204.07141 [cs, eess]},
}

@inproceedings{ermolov_whitening_2021,
	title = {Whitening for {Self}-{Supervised} {Representation} {Learning}},
	url = {https://proceedings.mlr.press/v139/ermolov21a.html},
	language = {en},
	urldate = {2024-05-22},
	booktitle = {Proceedings of the 38th {International} {Conference} on {Machine} {Learning}},
	publisher = {PMLR},
	author = {Ermolov, Aleksandr and Siarohin, Aliaksandr and Sangineto, Enver and Sebe, Nicu},
	month = jul,
	year = {2021},
	note = {ISSN: 2640-3498},
	pages = {3015--3024},
}

@inproceedings{xiao_what_2021,
	title = {What {Should} {Not} {Be} {Contrastive} in {Contrastive} {Learning}},
	url = {https://openreview.net/forum?id=CZ8Y3NzuVzO},
	language = {en},
	urldate = {2024-10-02},
	booktitle = {International {Conference} on {Learning} {Representations}},
	author = {Xiao, Tete and Wang, Xiaolong and Efros, Alexei A. and Darrell, Trevor},
	year = {2021},
}

@inproceedings{wang_understanding_2024,
	title = {Understanding the {Role} of {Equivariance} in {Self}-supervised {Learning}},
	url = {https://openreview.net/forum?id=NLqdudgBfy},
	language = {en},
	urldate = {2025-03-14},
	booktitle = {The {Thirty}-eighth {Annual} {Conference} on {Neural} {Information} {Processing} {Systems}},
	author = {Wang, Yifei and Hu, Kaiwen and Gupta, Sharut and Ye, Ziyu and Wang, Yisen and Jegelka, Stefanie},
	month = nov,
	year = {2024},
}

@article{mcnaughton_path_2006,
	title = {Path integration and the neural basis of the 'cognitive map'},
	volume = {7},
	copyright = {2006 Springer Nature Limited},
	issn = {1471-0048},
	url = {https://www.nature.com/articles/nrn1932},
	doi = {10.1038/nrn1932},
	language = {en},
	number = {8},
	urldate = {2025-04-09},
	journal = {Nature Reviews Neuroscience},
	author = {McNaughton, Bruce L. and Battaglia, Francesco P. and Jensen, Ole and Moser, Edvard I. and Moser, May-Britt},
	month = aug,
	year = {2006},
	note = {Publisher: Nature Publishing Group},
	pages = {663--678},
}

@inproceedings{haochen_provable_2021,
	title = {Provable {Guarantees} for {Self}-{Supervised} {Deep} {Learning} with {Spectral} {Contrastive} {Loss}},
	volume = {34},
	urldate = {2024-08-12},
	booktitle = {Advances in {Neural} {Information} {Processing} {Systems}},
	publisher = {Curran Associates, Inc.},
	author = {HaoChen, Jeff Z. and Wei, Colin and Gaidon, Adrien and Ma, Tengyu},
	year = {2021},
	pages = {5000--5011},
}

@inproceedings{lee_improving_2021,
	title = {Improving {Transferability} of {Representations} via {Augmentation}-{Aware} {Self}-{Supervision}},
	volume = {34},
	urldate = {2023-10-05},
	booktitle = {Advances in {Neural} {Information} {Processing} {Systems}},
	publisher = {Curran Associates, Inc.},
	author = {Lee, Hankook and Lee, Kibok and Lee, Kimin and Lee, Honglak and Shin, Jinwoo},
	year = {2021},
	pages = {17710--17722},
}

@inproceedings{vaswani_attention_2017,
	title = {Attention is {All} you {Need}},
	volume = {30},
	urldate = {2024-08-08},
	booktitle = {Advances in {Neural} {Information} {Processing} {Systems}},
	publisher = {Curran Associates, Inc.},
	author = {Vaswani, Ashish and Shazeer, Noam and Parmar, Niki and Uszkoreit, Jakob and Jones, Llion and Gomez, Aidan N and Kaiser, Lukasz and Polosukhin, Illia},
	year = {2017},
}

@inproceedings{shakerinava_structuring_2022,
	title = {Structuring {Representations} {Using} {Group} {Invariants}},
	volume = {35},
	language = {en},
	urldate = {2025-03-14},
	booktitle = {Advances in {Neural} {Information} {Processing} {Systems}},
	author = {Shakerinava, Mehran and Mondal, Arnab Kumar and Ravanbakhsh, Siamak},
	month = dec,
	year = {2022},
	pages = {34162--34174},
}

@inproceedings{guo_byol-explore_2022,
	title = {{BYOL}-{Explore}: {Exploration} by {Bootstrapped} {Prediction}},
	volume = {35},
	shorttitle = {{BYOL}-{Explore}},
	language = {en},
	urldate = {2024-09-08},
	booktitle = {Advances in {Neural} {Information} {Processing} {Systems}},
	author = {Guo, Zhaohan and Thakoor, Shantanu and Pislar, Miruna and Avila Pires, Bernardo and Altché, Florent and Tallec, Corentin and Saade, Alaa and Calandriello, Daniele and Grill, Jean-Bastien and Tang, Yunhao and Valko, Michal and Munos, Remi and Gheshlaghi Azar, Mohammad and Piot, Bilal},
	month = dec,
	year = {2022},
	pages = {31855--31870},
}

@inproceedings{grill_bootstrap_2020,
	title = {Bootstrap {Your} {Own} {Latent} - {A} {New} {Approach} to {Self}-{Supervised} {Learning}},
	volume = {33},
	urldate = {2024-08-12},
	booktitle = {Advances in {Neural} {Information} {Processing} {Systems}},
	publisher = {Curran Associates, Inc.},
	author = {Grill, Jean-Bastien and Strub, Florian and Altché, Florent and Tallec, Corentin and Richemond, Pierre and Buchatskaya, Elena and Doersch, Carl and Avila Pires, Bernardo and Guo, Zhaohan and Gheshlaghi Azar, Mohammad and Piot, Bilal and kavukcuoglu, koray and Munos, Remi and Valko, Michal},
	year = {2020},
	pages = {21271--21284},
}

@inproceedings{caron_unsupervised_2020,
	title = {Unsupervised {Learning} of {Visual} {Features} by {Contrasting} {Cluster} {Assignments}},
	volume = {33},
	urldate = {2023-11-16},
	booktitle = {Advances in {Neural} {Information} {Processing} {Systems}},
	publisher = {Curran Associates, Inc.},
	author = {Caron, Mathilde and Misra, Ishan and Mairal, Julien and Goyal, Priya and Bojanowski, Piotr and Joulin, Armand},
	year = {2020},
	pages = {9912--9924},
}

@inproceedings{dwibedi_little_2021,
	title = {With a {Little} {Help} {From} {My} {Friends}: {Nearest}-{Neighbor} {Contrastive} {Learning} of {Visual} {Representations}},
	shorttitle = {With a {Little} {Help} {From} {My} {Friends}},
	language = {en},
	urldate = {2024-08-12},
	booktitle = {Proceedings of the {IEEE}/{CVF} {International} {Conference} on {Computer} {Vision}},
	author = {Dwibedi, Debidatta and Aytar, Yusuf and Tompson, Jonathan and Sermanet, Pierre and Zisserman, Andrew},
	year = {2021},
	pages = {9588--9597},
}

@inproceedings{khetarpal_unifying_2025,
	title = {A {Unifying} {Framework} for {Action}-{Conditional} {Self}-{Predictive} {Reinforcement} {Learning}},
	url = {https://openreview.net/forum?id=5DypCUsMg4},
	language = {en},
	urldate = {2025-03-14},
	booktitle = {The 28th {International} {Conference} on {Artificial} {Intelligence} and {Statistics}},
	author = {Khetarpal, Khimya and Guo, Zhaohan Daniel and Pires, Bernardo Avila and Tang, Yunhao and Lyle, Clare and Rowland, Mark and Heess, Nicolas and Borsa, Diana L. and Guez, Arthur and Dabney, Will},
	month = feb,
	year = {2025},
}

@inproceedings{scherr_self-supervised_2022,
	title = {Self-{Supervised} {Learning} {Through} {Efference} {Copies}},
	url = {https://openreview.net/forum?id=DotEQCtY67g},
	language = {en},
	urldate = {2025-03-14},
	booktitle = {Advances in {Neural} {Information} {Processing} {Systems}},
	author = {Scherr, Franz and Guo, Qinghai and Moraitis, Timoleon},
	month = oct,
	year = {2022},
}

@inproceedings{ni_bridging_2024,
	title = {Bridging {State} and {History} {Representations}: {Understanding} {Self}-{Predictive} {RL}},
	shorttitle = {Bridging {State} and {History} {Representations}},
	url = {https://openreview.net/forum?id=ms0VgzSGF2},
	language = {en},
	urldate = {2025-01-06},
	booktitle = {The {Twelfth} {International} {Conference} on {Learning} {Representations}},
	author = {Ni, Tianwei and Eysenbach, Benjamin and SeyedSalehi, Erfan and Ma, Michel and Gehring, Clement and Mahajan, Aditya and Bacon, Pierre-Luc},
	year = {2024},
}

@misc{sobal_learning_2025,
	title = {Learning from {Reward}-{Free} {Offline} {Data}: {A} {Case} for {Planning} with {Latent} {Dynamics} {Models}},
	shorttitle = {Learning from {Reward}-{Free} {Offline} {Data}},
	url = {http://arxiv.org/abs/2502.14819},
	doi = {10.48550/arXiv.2502.14819},
	urldate = {2025-03-06},
	publisher = {arXiv},
	author = {Sobal, Vlad and Zhang, Wancong and Cho, Kynghyun and Balestriero, Randall and Rudner, Tim G. J. and LeCun, Yann},
	month = feb,
	year = {2025},
	note = {arXiv:2502.14819 [cs]},
}

@inproceedings{yeh_decoupled_2022,
	address = {Cham},
	title = {Decoupled {Contrastive} {Learning}},
	isbn = {978-3-031-19809-0},
	doi = {10.1007/978-3-031-19809-0_38},
	language = {en},
	booktitle = {Computer {Vision} – {ECCV} 2022},
	publisher = {Springer Nature Switzerland},
	author = {Yeh, Chun-Hsiao and Hong, Cheng-Yao and Hsu, Yen-Chi and Liu, Tyng-Luh and Chen, Yubei and LeCun, Yann},
	editor = {Avidan, Shai and Brostow, Gabriel and Cissé, Moustapha and Farinella, Giovanni Maria and Hassner, Tal},
	year = {2022},
	pages = {668--684},
}

@inproceedings{song_ego4d_2023,
	title = {{Ego4D} {Goal}-{Step}: {Toward} {Hierarchical} {Understanding} of {Procedural} {Activities}},
	shorttitle = {{Ego4D} {Goal}-{Step}},
	url = {https://openreview.net/forum?id=3BxYAaovKr&noteId=IoK9NhxnlM},
	language = {en},
	urldate = {2025-02-10},
	author = {Song, Yale and Byrne, Gene and Nagarajan, Tushar and Wang, Huiyu and Martin, Miguel and Torresani, Lorenzo},
	month = nov,
	year = {2023},
}

@inproceedings{bardes_vicreg_2022,
	title = {{VICReg}: {Variance}-{Invariance}-{Covariance} {Regularization} for {Self}-{Supervised} {Learning}},
	shorttitle = {{VICReg}},
	url = {https://openreview.net/forum?id=xm6YD62D1Ub},
	language = {en},
	urldate = {2024-08-13},
	booktitle = {International {Conference} on {Learning} {Representations}},
	author = {Bardes, Adrien and Ponce, Jean and LeCun, Yann},
	year = {2022},
}

@misc{lin_vedit_2024,
	title = {{VEDIT}: {Latent} {Prediction} {Architecture} {For} {Procedural} {Video} {Representation} {Learning}},
	shorttitle = {{VEDIT}},
	url = {http://arxiv.org/abs/2410.03478},
	doi = {10.48550/arXiv.2410.03478},
	urldate = {2025-01-05},
	publisher = {arXiv},
	author = {Lin, Han and Nagarajan, Tushar and Ballas, Nicolas and Assran, Mido and Komeili, Mojtaba and Bansal, Mohit and Sinha, Koustuv},
	month = oct,
	year = {2024},
	note = {arXiv:2410.03478 [cs]},
}

@inproceedings{gupta_-context_2024,
	title = {In-{Context} {Symmetries}: {Self}-{Supervised} {Learning} through {Contextual} {World} {Models}},
	shorttitle = {In-{Context} {Symmetries}},
	url = {https://openreview.net/forum?id=etPAH4xSUn},
	language = {en},
	urldate = {2025-01-03},
	booktitle = {The {Thirty}-eighth {Annual} {Conference} on {Neural} {Information} {Processing} {Systems}},
	author = {Gupta, Sharut and Wang, Chenyu and Wang, Yifei and Jaakkola, Tommi and Jegelka, Stefanie},
	month = nov,
	year = {2024},
}

@inproceedings{yerxa_contrastive-equivariant_2024,
	title = {Contrastive-{Equivariant} {Self}-{Supervised} {Learning} {Improves} {Alignment} with {Primate} {Visual} {Area} {IT}},
	url = {https://openreview.net/forum?id=AiMs8GPP5q},
	language = {en},
	urldate = {2024-11-16},
	author = {Yerxa, Thomas Edward and Feather, Jenelle and Simoncelli, Eero P. and Chung, SueYeon},
	month = nov,
	year = {2024},
}

@inproceedings{schwarzer_data-efficient_2021,
	title = {Data-{Efficient} {Reinforcement} {Learning} with {Self}-{Predictive} {Representations}},
	url = {https://openreview.net/forum?id=uCQfPZwRaUu&fbclid=IwAR3FMvlynXXYEMJaJzPki1x1wC9jjA3aBDC_moWxrI91hLaDvtk7nnnIXT8},
	language = {en},
	urldate = {2024-10-02},
	booktitle = {International {Conference} on {Learning} {Representations}},
	author = {Schwarzer, Max and Anand, Ankesh and Goel, Rishab and Hjelm, R. Devon and Courville, Aaron and Bachman, Philip},
	year = {2021},
}

@inproceedings{tang_understanding_2023,
	title = {Understanding {Self}-{Predictive} {Learning} for {Reinforcement} {Learning}},
	url = {https://proceedings.mlr.press/v202/tang23d.html},
	language = {en},
	urldate = {2024-10-18},
	booktitle = {Proceedings of the 40th {International} {Conference} on {Machine} {Learning}},
	publisher = {PMLR},
	author = {Tang, Yunhao and Guo, Zhaohan Daniel and Richemond, Pierre Harvey and Pires, Bernardo Avila and Chandak, Yash and Munos, Remi and Rowland, Mark and Azar, Mohammad Gheshlaghi and Lan, Charline Le and Lyle, Clare and György, András and Thakoor, Shantanu and Dabney, Will and Piot, Bilal and Calandriello, Daniele and Valko, Michal},
	month = jul,
	year = {2023},
	note = {ISSN: 2640-3498},
	pages = {33632--33656},
}

@inproceedings{guo_bootstrap_2020,
	title = {Bootstrap {Latent}-{Predictive} {Representations} for {Multitask} {Reinforcement} {Learning}},
	url = {https://proceedings.mlr.press/v119/guo20g.html},
	language = {en},
	urldate = {2024-10-08},
	booktitle = {Proceedings of the 37th {International} {Conference} on {Machine} {Learning}},
	publisher = {PMLR},
	author = {Guo, Zhaohan Daniel and Pires, Bernardo Avila and Piot, Bilal and Grill, Jean-Bastien and Altché, Florent and Munos, Remi and Azar, Mohammad Gheshlaghi},
	month = nov,
	year = {2020},
	note = {ISSN: 2640-3498},
	pages = {3875--3886},
}

@article{posner_inhibition_1985,
	title = {Inhibition of return: {Neural} basis and function},
	volume = {2},
	issn = {0264-3294, 1464-0627},
	shorttitle = {Inhibition of return},
	url = {http://www.tandfonline.com/doi/abs/10.1080/02643298508252866},
	doi = {10.1080/02643298508252866},
	language = {en},
	number = {3},
	urldate = {2024-11-03},
	journal = {Cognitive Neuropsychology},
	author = {Posner, Michael I. and Rafal, Robert D. and Choate, Lisa S. and Vaughan, Jonathan},
	month = aug,
	year = {1985},
	pages = {211--228},
}

@misc{kipf_contrastive_2020,
	title = {Contrastive {Learning} of {Structured} {World} {Models}},
	url = {http://arxiv.org/abs/1911.12247},
	language = {en},
	urldate = {2024-05-21},
	publisher = {arXiv},
	author = {Kipf, Thomas and van der Pol, Elise and Welling, Max},
	month = jan,
	year = {2020},
	note = {arXiv:1911.12247 [cs, stat]},
}

@inproceedings{gidaris_unsupervised_2018,
	title = {Unsupervised {Representation} {Learning} by {Predicting} {Image} {Rotations}},
	url = {https://openreview.net/forum?id=S1v4N2l0-},
	language = {en},
	urldate = {2024-08-14},
	author = {Gidaris, Spyros and Singh, Praveer and Komodakis, Nikos},
	month = feb,
	year = {2018},
}

@misc{garrido_learning_2024,
	title = {Learning and {Leveraging} {World} {Models} in {Visual} {Representation} {Learning}},
	url = {http://arxiv.org/abs/2403.00504},
	language = {en},
	urldate = {2024-04-26},
	publisher = {arXiv},
	author = {Garrido, Quentin and Assran, Mahmoud and Ballas, Nicolas and Bardes, Adrien and Najman, Laurent and LeCun, Yann},
	month = mar,
	year = {2024},
	note = {arXiv:2403.00504 [cs]},
}

@article{itti_model_1998,
	title = {A model of saliency-based visual attention for rapid scene analysis},
	volume = {20},
	issn = {1939-3539},
	url = {https://ieeexplore.ieee.org/document/730558/?arnumber=730558},
	doi = {10.1109/34.730558},
	number = {11},
	urldate = {2024-10-20},
	journal = {IEEE Transactions on Pattern Analysis and Machine Intelligence},
	author = {Itti, L. and Koch, C. and Niebur, E.},
	month = nov,
	year = {1998},
	note = {Conference Name: IEEE Transactions on Pattern Analysis and Machine Intelligence},
	pages = {1254--1259},
}

@inproceedings{wang2024pose,
  title={Pose-aware self-supervised learning with viewpoint trajectory regularization},
  author={Wang, Jiayun and Chen, Yubei and Yu, Stella X},
  booktitle={European Conference on Computer Vision},
  pages={19--37},
  year={2024},
  organization={Springer}
}

@inproceedings{gupta_structuring_2023,
	title = {Structuring {Representation} {Geometry} with {Rotationally} {Equivariant} {Contrastive} {Learning}},
	url = {https://openreview.net/forum?id=lgaFMvZHSJ},
	language = {en},
	urldate = {2024-08-14},
	booktitle = {The {Twelfth} {International} {Conference} on {Learning} {Representations}},
	author = {Gupta, Sharut and Robinson, Joshua and Lim, Derek and Villar, Soledad and Jegelka, Stefanie},
	month = oct,
	year = {2023},
}

@inproceedings{devillers_equimod_2022,
	title = {{EquiMod}: {An} {Equivariance} {Module} to {Improve} {Visual} {Instance} {Discrimination}},
	shorttitle = {{EquiMod}},
	url = {https://openreview.net/forum?id=eDLwjKmtYFt},
	language = {en},
	urldate = {2024-08-12},
	booktitle = {The {Eleventh} {International} {Conference} on {Learning} {Representations}},
	author = {Devillers, Alexandre and Lefort, Mathieu},
	month = sep,
	year = {2022},
}

@article{vuilleumier_multiple_2002,
	title = {Multiple levels of visual object constancy revealed by event-related {fMRI} of repetition priming},
	volume = {5},
	copyright = {2002 Springer Nature America, Inc.},
	issn = {1546-1726},
	url = {https://www.nature.com/articles/nn839},
	doi = {10.1038/nn839},
	language = {en},
	number = {5},
	urldate = {2024-09-19},
	journal = {Nature Neuroscience},
	author = {Vuilleumier, P. and Henson, R. N. and Driver, J. and Dolan, R. J.},
	month = may,
	year = {2002},
	note = {Publisher: Nature Publishing Group},
	pages = {491--499},
}

@article{harman_active_1999,
	title = {Active manual control of object views facilitates visual recognition},
	volume = {9},
	copyright = {https://www.elsevier.com/tdm/userlicense/1.0/},
	issn = {09609822},
	url = {https://linkinghub.elsevier.com/retrieve/pii/S0960982200800536},
	doi = {10.1016/S0960-9822(00)80053-6},
	language = {en},
	number = {22},
	urldate = {2024-09-19},
	journal = {Current Biology},
	author = {Harman, Karin L. and Humphrey, G.Keith and Goodale, Melvyn A.},
	month = nov,
	year = {1999},
	pages = {1315--1318},
}

@article{tarr_three-dimensional_1998,
	title = {Three-dimensional object recognition is viewpoint dependent},
	volume = {1},
	copyright = {1998 Nature America Inc.},
	issn = {1546-1726},
	url = {https://www.nature.com/articles/nn0898_275},
	doi = {10.1038/1089},
	language = {en},
	number = {4},
	urldate = {2024-09-19},
	journal = {Nature Neuroscience},
	author = {Tarr, Michael J. and Williams, Pepper and Hayward, William G. and Gauthier, Isabel},
	month = aug,
	year = {1998},
	note = {Publisher: Nature Publishing Group},
	pages = {275--277},
}

@misc{van_den_oord_representation_2019,
	title = {Representation {Learning} with {Contrastive} {Predictive} {Coding}},
	url = {http://arxiv.org/abs/1807.03748},
	doi = {10.48550/arXiv.1807.03748},
	urldate = {2023-09-22},
	publisher = {arXiv},
	author = {van den Oord, Aaron and Li, Yazhe and Vinyals, Oriol},
	month = jan,
	year = {2019},
	note = {arXiv:1807.03748 [cs, stat]},
}

@inproceedings{misra_self-supervised_2020,
	title = {Self-{Supervised} {Learning} of {Pretext}-{Invariant} {Representations}},
	url = {https://openaccess.thecvf.com/content_CVPR_2020/html/Misra_Self-Supervised_Learning_of_Pretext-Invariant_Representations_CVPR_2020_paper.html},
	urldate = {2024-08-12},
	booktitle = {Proceedings of the {IEEE}/{CVF} {Conference} on {Computer} {Vision} and {Pattern} {Recognition}},
	author = {Misra, Ishan and van der Maaten, Laurens},
	year = {2020},
	pages = {6707--6717},
}

@inproceedings{dangovski_equivariant_2022,
	title = {Equivariant {Self}-{Supervised} {Learning}: {Encouraging} {Equivariance} in {Representations}},
	shorttitle = {Equivariant {Self}-{Supervised} {Learning}},
	url = {https://openreview.net/forum?id=gKLAAfiytI},
	language = {en},
	urldate = {2024-08-12},
	booktitle = {International {Conference} on {Learning} {Representations}},
	author = {Dangovski, Rumen and Jing, Li and Loh, Charlotte and Han, Seungwook and Srivastava, Akash and Cheung, Brian and Agrawal, Pulkit and Soljacic, Marin},
	year = {2022},
}

@inproceedings{he_deep_2016,
	address = {Las Vegas, NV, USA},
	title = {Deep {Residual} {Learning} for {Image} {Recognition}},
	isbn = {978-1-4673-8851-1},
	url = {http://ieeexplore.ieee.org/document/7780459/},
	doi = {10.1109/CVPR.2016.90},
	language = {en},
	urldate = {2024-08-13},
	booktitle = {2016 {IEEE} {Conference} on {Computer} {Vision} and {Pattern} {Recognition} ({CVPR})},
	publisher = {IEEE},
	author = {He, Kaiming and Zhang, Xiangyu and Ren, Shaoqing and Sun, Jian},
	month = jun,
	year = {2016},
	pages = {770--778},
}

@inproceedings{he_momentum_2020,
	address = {Seattle, WA, USA},
	title = {Momentum {Contrast} for {Unsupervised} {Visual} {Representation} {Learning}},
	copyright = {https://ieeexplore.ieee.org/Xplorehelp/downloads/license-information/IEEE.html},
	isbn = {978-1-72817-168-5},
	url = {https://ieeexplore.ieee.org/document/9157636/},
	doi = {10.1109/CVPR42600.2020.00975},
	urldate = {2024-08-12},
	booktitle = {2020 {IEEE}/{CVF} {Conference} on {Computer} {Vision} and {Pattern} {Recognition} ({CVPR})},
	publisher = {IEEE},
	author = {He, Kaiming and Fan, Haoqi and Wu, Yuxin and Xie, Saining and Girshick, Ross},
	month = jun,
	year = {2020},
	pages = {9726--9735},
}

@inproceedings{dosovitskiy_image_2020,
	title = {An {Image} is {Worth} 16x16 {Words}: {Transformers} for {Image} {Recognition} at {Scale}},
	shorttitle = {An {Image} is {Worth} 16x16 {Words}},
	url = {https://openreview.net/forum?id=YicbFdNTTy},
	language = {en},
	urldate = {2024-08-12},
	booktitle = {International {Conference} on {Learning} {Representations}},
	author = {Dosovitskiy, Alexey and Beyer, Lucas and Kolesnikov, Alexander and Weissenborn, Dirk and Zhai, Xiaohua and Unterthiner, Thomas and Dehghani, Mostafa and Minderer, Matthias and Heigold, Georg and Gelly, Sylvain and Uszkoreit, Jakob and Houlsby, Neil},
	month = oct,
	year = {2020},
}

@misc{lecun_path_2022,
	type = {{OpenReview}},
	title = {A {Path} {Towards} {Autonomous} {Machine} {Intelligence} {Version} 0.9.2, 2022-06-27},
	language = {en},
	publisher = {OpenReview},
	author = {LeCun, Yann},
	year = {2022},
	note = {OpenReview},
}

@inproceedings{park_learning_2022,
	title = {Learning {Symmetric} {Embeddings} for {Equivariant} {World} {Models}},
	url = {https://proceedings.mlr.press/v162/park22a.html},
	language = {en},
	urldate = {2024-08-12},
	booktitle = {Proceedings of the 39th {International} {Conference} on {Machine} {Learning}},
	publisher = {PMLR},
	author = {Park, Jung Yeon and Biza, Ondrej and Zhao, Linfeng and Meent, Jan-Willem Van De and Walters, Robin},
	month = jun,
	year = {2022},
	note = {ISSN: 2640-3498},
	pages = {17372--17389},
}

@inproceedings{brown_language_2020,
	title = {Language {Models} are {Few}-{Shot} {Learners}},
	volume = {33},
	url = {https://proceedings.neurips.cc/paper_files/paper/2020/hash/1457c0d6bfcb4967418bfb8ac142f64a-Abstract.html},
	urldate = {2024-08-08},
	booktitle = {Advances in {Neural} {Information} {Processing} {Systems}},
	publisher = {Curran Associates, Inc.},
	author = {Brown, Tom and Mann, Benjamin and Ryder, Nick and Subbiah, Melanie and Kaplan, Jared D and Dhariwal, Prafulla and Neelakantan, Arvind and Shyam, Pranav and Sastry, Girish and Askell, Amanda and Agarwal, Sandhini and Herbert-Voss, Ariel and Krueger, Gretchen and Henighan, Tom and Child, Rewon and Ramesh, Aditya and Ziegler, Daniel and Wu, Jeffrey and Winter, Clemens and Hesse, Chris and Chen, Mark and Sigler, Eric and Litwin, Mateusz and Gray, Scott and Chess, Benjamin and Clark, Jack and Berner, Christopher and McCandlish, Sam and Radford, Alec and Sutskever, Ilya and Amodei, Dario},
	year = {2020},
	pages = {1877--1901},
}

@inproceedings{zbontar_barlow_2021,
	title = {Barlow {Twins}: {Self}-{Supervised} {Learning} via {Redundancy} {Reduction}},
	shorttitle = {Barlow {Twins}},
	url = {https://proceedings.mlr.press/v139/zbontar21a.html},
	language = {en},
	urldate = {2024-08-12},
	booktitle = {Proceedings of the 38th {International} {Conference} on {Machine} {Learning}},
	publisher = {PMLR},
	author = {Zbontar, Jure and Jing, Li and Misra, Ishan and LeCun, Yann and Deny, Stephane},
	month = jul,
	year = {2021},
	note = {ISSN: 2640-3498},
	pages = {12310--12320},
}

@inproceedings{garrido_self-supervised_2023,
	title = {Self-supervised learning of {Split} {Invariant} {Equivariant} representations},
	url = {https://proceedings.mlr.press/v202/garrido23b.html},
	language = {en},
	urldate = {2024-08-12},
	booktitle = {Proceedings of the 40th {International} {Conference} on {Machine} {Learning}},
	publisher = {PMLR},
	author = {Garrido, Quentin and Najman, Laurent and Lecun, Yann},
	month = jul,
	year = {2023},
	note = {ISSN: 2640-3498},
	pages = {10975--10996},
}

@inproceedings{chen_exploring_2021,
	title = {Exploring {Simple} {Siamese} {Representation} {Learning}},
	url = {https://openaccess.thecvf.com/content/CVPR2021/html/Chen_Exploring_Simple_Siamese_Representation_Learning_CVPR_2021_paper.html},
	language = {en},
	urldate = {2024-08-12},
	booktitle = {Proceedings of the {IEEE}/{CVF} {Conference} on {Computer} {Vision} and {Pattern} {Recognition}},
	author = {Chen, Xinlei and He, Kaiming},
	year = {2021},
	pages = {15750--15758},
}

@inproceedings{chen_simple_2020,
	title = {A {Simple} {Framework} for {Contrastive} {Learning} of {Visual} {Representations}},
	url = {https://proceedings.mlr.press/v119/chen20j.html},
	language = {en},
	urldate = {2024-08-12},
	booktitle = {Proceedings of the 37th {International} {Conference} on {Machine} {Learning}},
	publisher = {PMLR},
	author = {Chen, Ting and Kornblith, Simon and Norouzi, Mohammad and Hinton, Geoffrey},
	month = nov,
	year = {2020},
	note = {ISSN: 2640-3498},
	pages = {1597--1607},
}

@inproceedings{caron_emerging_2021,
	title = {Emerging {Properties} in {Self}-{Supervised} {Vision} {Transformers}},
	url = {https://openaccess.thecvf.com/content/ICCV2021/html/Caron_Emerging_Properties_in_Self-Supervised_Vision_Transformers_ICCV_2021_paper},
	language = {en},
	urldate = {2024-08-12},
	booktitle = {Proceedings of the {IEEE}/{CVF} {International} {Conference} on {Computer} {Vision}},
	author = {Caron, Mathilde and Touvron, Hugo and Misra, Ishan and Jégou, Hervé and Mairal, Julien and Bojanowski, Piotr and Joulin, Armand},
	year = {2021},
	pages = {9650--9660},
}

@inproceedings{assran_self-supervised_2023,
	address = {Vancouver, BC, Canada},
	title = {Self-{Supervised} {Learning} from {Images} with a {Joint}-{Embedding} {Predictive} {Architecture}},
	copyright = {https://doi.org/10.15223/policy-029},
	isbn = {979-8-3503-0129-8},
	url = {https://ieeexplore.ieee.org/document/10205476/},
	doi = {10.1109/CVPR52729.2023.01499},
	language = {en},
	urldate = {2024-08-12},
	booktitle = {2023 {IEEE}/{CVF} {Conference} on {Computer} {Vision} and {Pattern} {Recognition} ({CVPR})},
	publisher = {IEEE},
	author = {Assran, Mahmoud and Duval, Quentin and Misra, Ishan and Bojanowski, Piotr and Vincent, Pascal and Rabbat, Michael and LeCun, Yann and Ballas, Nicolas},
	month = jun,
	year = {2023},
	pages = {15619--15629},
}

@misc{chen_empirical_2021,
	title = {An {Empirical} {Study} of {Training} {Self}-{Supervised} {Vision} {Transformers}},
	url = {http://arxiv.org/abs/2104.02057},
	doi = {10.48550/arXiv.2104.02057},
	urldate = {2024-06-14},
	publisher = {arXiv},
	author = {Chen, Xinlei and Xie, Saining and He, Kaiming},
	month = aug,
	year = {2021},
	note = {arXiv:2104.02057 [cs]},
}

@incollection{zhaoping_v1_2014,
	edition = {1},
	title = {The {V1} hypothesis—creating a bottom-up saliency map for preattentive selection and segmentation},
	isbn = {978-0-19-956466-8 978-0-19-177250-4},
	url = {https://academic.oup.com/book/8719/chapter/154784147},
	language = {en},
	urldate = {2024-05-15},
	booktitle = {Understanding {Vision}},
	publisher = {Oxford University PressOxford},
	author = {Zhaoping, Li},
	collaborator = {Zhaoping, Li},
	month = may,
	year = {2014},
	doi = {10.1093/acprof:oso/9780199564668.003.0005},
	doi = {10.1093/acprof:oso/9780199564668.003.0005},
	pages = {189--314},
}

@article{li_saliency_2002,
	title = {A saliency map in primary visual cortex},
	volume = {6},
	issn = {1364-6613, 1879-307X},
	url = {https://www.cell.com/trends/cognitive-sciences/abstract/S1364-6613(00)01817-9},
	doi = {10.1016/S1364-6613(00)01817-9},
	language = {English},
	number = {1},
	urldate = {2024-05-15},
	journal = {Trends in Cognitive Sciences},
	author = {Li, Zhaoping},
	month = jan,
	year = {2002},
	pmid = {11849610},
	note = {Publisher: Elsevier},
	pages = {9--16},
}

@inproceedings{linardos_deepgaze_2021,
	address = {Montreal, QC, Canada},
	title = {{DeepGaze} {IIE}: {Calibrated} prediction in and out-of-domain for state-of-the-art saliency modeling},
	isbn = {978-1-6654-2812-5},
	shorttitle = {{DeepGaze} {IIE}},
	url = {https://ieeexplore.ieee.org/document/9711473/},
	doi = {10.1109/ICCV48922.2021.01268},
	language = {en},
	urldate = {2023-11-01},
	booktitle = {2021 {IEEE}/{CVF} {International} {Conference} on {Computer} {Vision} ({ICCV})},
	publisher = {IEEE},
	author = {Linardos, Akis and Kummerer, Matthias and Press, Ori and Bethge, Matthias},
	month = oct,
	year = {2021},
	pages = {12899--12908},
}
\bibliographystyle{plainnat}







\appendix

\section{Implementation Details}\label{app1}

\subsection{Data preparation}\label{sec:dataprep}

\textbf{3DIEBench.} The original $256\times 256$ images are resized to a $128\times 128$ resolution for all experiments. Normalization is done using the means and standard deviations in \citet{garrido_self-supervised_2023}, i.e., $\mu=[0.5016, 0.5037, 0.5060]$ and $\sigma=[0.1030, 0.0999, 0.0969]$ for the three RGB channels, respectively.

\textbf{CIFAR100.} We use $32\times 32$ images with normalization parameters typically used in the literature, i.e., $\mu=[0.4914, 0.4822, 0.4465]$ and $\sigma=[0.247, 0.243, 0.261]$. For data augmentation, we follow EquiMod's augmentation strategy. 

\textbf{Tiny ImageNet.} The training set consists of 100000 ImageNet-1k images from 200 classes (500 for each class) downsized to $64\times 64$. The validation set has 50 images per class. We use normalization parameters $\mu=[0.4914, 0.4822, 0.4465]$ and $\sigma=[0.247, 0.243, 0.261]$, for the three RGB channels, respectively. For data augmentation, we use the same augmentation parameters as CIFAR100 with the kernel size of Gaussian blur adapted to the $64\times 64$ images.

\textbf{STL10.} In order to extract the saliencies, we resize images to $512\times 512$, feed them to the pre-trained DeepGaze IIE \citep{linardos_deepgaze_2021}, resize the output saliencies back to $96\times 96$, and store them alongside original images. We use normalization parameters $\mu=[0.4467, 0.4398, 0.4066]$, $\sigma=[0.2241, 0.2215, 0.2239]$ for the three RGB channels, respectively. After sampling fixations from saliencies, the patches that are extracted from the image to simulate foveation are $32\times 32$ (compared to the full image size of $96\times 96$). For IoR, we zero-out a circular area with radius of $16$ around each previous fixation.

\textbf{Transformation parameters.} For 3DIEBench, we use the rotation and color parameters provided with images for action conditioning as done in \citet{garrido_self-supervised_2023}. For the augmentation setting, we use the parameters corresponding to each of the three augmentations and form the action as the relative augmentation vector between two images. For crop, we use four variables, i.e., vertical and horizontal coordinate, and height and width. For color jitter, we use four variables: brightness, contrast, saturation, and hue. For blur, we use one variable: the standard deviation of the blurring kernel. In predictive learning across saccades, the action is a 2-d vector, i.e., the normalized relative $(x,y)$ coordinate between two patches.

\subsection{Training and evaluation details}\label{sec:evaldetails}

\paragraph{Additional Training Details.} We used the PyTorch framework for training all models.  For experiments that use CIFAR100 and low-resolution STL-10 patches in predictive learning across saccades, we use the CIFAR variant of ResNet-18. For models trained with AdamW, we used with default $\beta_1$ and $\beta_2$, a weight decay of $0.001$, and a learning rate of $4\times 10^{-4}$ with a linear warmup for $20$ epochs starting from $10^{-5}$ followed by a cosine decay back to $10^{-5}$. For models trained with Adam, we use the Adam optimizer with a learning rate of $10^{-3}$, default $\beta_1$ and $\beta_2$, and no weight decay.
 
\paragraph{Protocols for training evaluation heads.} For linear probing, we follow a common SSL protocol and train a linear classifier on top of frozen representations with a batch size of 256 for 300 epochs using the Adam optimizer with default hyperparameters. For action prediction, we follow a similar protocol as SIE \citep{garrido_self-supervised_2023}. Specifically, for rotation, color jitter, and crop, we train an MLP regressor with a hidden dimension of 1024 and ReLU activation for 300 epochs. For color (in 3DIEBench), blur (in the augmentation setting), and position (in predictive learning across saccades), we use a linear regressor and train it for 50 epochs. For path integration experiments, the same regressor architectures as the equivariance evaluation heads are used, i.e., the MLP for angular path integration and the linear regressor for saccade path integration. All regression heads are trained using the Adam optimizer with default hyperparameters.

\textbf{Hardware.} Each experiment was run on a single NVIDIA A100 GPU with 40GB of accelerator RAM.

\textbf{Compute cost and FLOP analysis.} We report pretraining compute time across methods (see Table~\ref{tab:gpu_hours}). All experiments are run under the same configuration: a single A100 GPU, 128$\times$128 resolution, batch size 512, for 1000 epochs on 3DIEBench. seq-JEPA with a sequence length of one has similar runtime to other baselines (e.g., BYOL, SimCLR), while seq-JEPA with sequence length of three incurs a moderate increase in wall-clock time (15.1 GPU-hours) due to encoding additional action-view pairs.

\begin{table}[h!]
\centering
\caption{GPU-Hour comparison across methods}
\label{tab:gpu_hours}
\begin{threeparttable}
\begin{tabular}{l c}
\toprule
Method & A100 pretraining GPU hours \\
\midrule
SimCLR & 11.4 \\
BYOL & 11.1 \\
SIE & 12.0 \\
VICReg & 10.9 \\
EquiMod & 11.8 \\
seq-JEPA (train seq len = 1) & 12.3 \\
seq-JEPA (train seq len = 3) & 15.1 \\
\bottomrule
\end{tabular}
\end{threeparttable}
\end{table}

seq-JEPA's inference cost is primarily governed by two factors: sequence length and input resolution. Below, we report detailed FLOP counts (in Gigaflops) per forward pass of a single datapoint across a range of configurations (Note: these reflect inference-time cost).

\begin{table}[h!]
\centering \tiny
\caption{FLOPs for different seq-JEPA configurations (in gigaflops)}
\label{tab:flops}
\begin{threeparttable}
\renewcommand{\arraystretch}{1.1}
\begin{tabular}{lcccccc}
\toprule
\textbf{Resolution / Config} & \textbf{Encoder only} &
\multicolumn{5}{c}{\textbf{Post-aggregator}} \\
\cmidrule(lr){3-7}
 & & seq len = 1 & seq len = 2 & seq len = 3 & seq len = 4 & seq len = 8 \\
\midrule
Each view is 128$\times$128 & 0.60G & 0.63G & 1.24G & 1.85G & 2.46G & 4.9G \\
Views of 64$\times$64 vs. full image (224$\times$224) &
1.82G (224) / 0.15G (64) & 0.18G & 0.34G & 0.51G & 0.67G & 1.32G \\
Views of 84$\times$84, vs. full image (224$\times$224) &
1.82G (224) / 0.30G (84) & 0.33G & 0.64G & 0.95G & 1.26G & 2.51G \\
\bottomrule
\end{tabular}
\end{threeparttable}
\end{table}

We observe that for single-view inference at 128$\times$128 resolution, the encoder requires only 0.60G FLOPs, which matches the compute cost of standard SSL baselines such as SimCLR and BYOL. As sequence length increases, the additional compute cost grows sub-linearly as additional views should be encoded and aggregated. For example, lengths 2 and 3---where invariant performance improves substantially---incur $\sim$1.2G and $\sim$1.85G FLOPs, respectively. The cost can be offset by reducing input resolution. In our saccade-based setup, we can feed low-resolution foveal glimpses (e.g., 64$\times$64 or 84$\times$84 crops sampled via saliency maps), which require only 0.15G–0.30G FLOPs per view. This enables the use of longer sequences at an overall compute cost comparable to full-resolution, single-view pipelines (e.g., compare the FLOPs of aggregating eight 64$\times$64 crops with encoding a single 224$\times$224 image in the second row of Table~\ref{tab:flops}).

\subsection{Architectural details}\label{sec:archdetails}

Below, we describe the architectural details and hyperparameters specific to each baseline.

\paragraph{BYOL.} We use a a projection head with 2048-2048-2048 intermediate dimensions. The predictor has a hidden dimension of 512-d with ReLU activation. We use the same EMA setup outlined in the original paper \citep{guo_bootstrap_2020}, i.e., the EMA parameter $\tau$ starts from $\tau_{\text{base}} = 0.996$ and is increased following a cosine schedule.

\paragraph{SimCLR.} We use a temperature parameter of $\tau=0.5$ with a projection MLP with 2048-2048-2048 intermediate dimensions.

\paragraph{VICReg.} We use $\lambda_{\text{inv}} = \lambda_{V} = 10$, $\lambda_{C} = 1$, and a projection head with 2048-2048-2048 intermediate dimensions.

\paragraph{VICRegTraj.} We use the same architecture as VICReg and add a trajectory loss (Equation 2 in~\citet{wang2024pose}) with a coefficient $\lambda_{traj}=0.01$.

\paragraph{Conditional BYOL.} The architecture is the same as BYOL, except that the predictor also receives the normalized relative transformation parameters. We use a linear action projector of 128-d and the same EMA setup as BYOL.

\paragraph{SIE.} For both invariant and equivariant projection heads, we use intermediate dimensions of 1024-1024-1024. For the loss coefficients, we use $\lambda_{\text{inv}} = \lambda_{V} = 10$, $\lambda_{\text{equi}} = 4.5$, and $\lambda_{C} = 1$. We use the hypernetwork architecture for all experiments.

\paragraph{SEN.} We use a temperature parameter of $\tau=0.5$ with a projection MLP with 2048-2048-2048 intermediate dimensions.

\paragraph{EquiMod.} We use the version based on SimCLR (both invariant and equivariant losses are contrastive with $\tau=0.1$ and have equal weights). The projection head has 1024-1024-128 intermediate dimensions. We use a linear action projector of 128-d.

\paragraph{ContextSSL.} We use the pre-trained weights provided by the authors (trained for 1000 epochs) and follow their evaluation protocol on 3DIEBench.

\paragraph{seq-JEPA.} For the sequence aggregator, we use a transformer encoder with three layers, four attention heads, and post-normalization. For the predictor, we use an MLP with a hidden layer of 1024-d and ReLU activation. The linear action projector in our default setting is 128-d. We use the same EMA setup as BYOL.

\section{Additional Experimental Results}\label{app2}

\subsection{Evaluation results on 3DIEBench for models conditioned on rotation and color} \label{sec:rotcol}

Table~\ref{tab:3db_rot_color} reports performance of seq-JEPA and several equivariant baselines when conditioned on both rotation and color in 3DIEBench. All methods suffer a drop in classification performance and become highly sensitive to color. Similar performance degradations have been previously observed with 3DIEBench \citep{garrido_self-supervised_2023,gupta_-context_2024}, though without a clear explanation. One plausible explanation aligned with Principle II in~\citet{wang_understanding_2024}, is that color in 3DIEBench, i.e., floor and light hue, are weakly correlated with class labels in 3DIEBench (low class relevance). Therefore, forcing the encoder to encode color information would cause class information to be lost, resulting in degradation of classification accuracy.

\begin{table}[!h]
    \caption{\small Evaluation on 3DIEBench for rotation and color prediction (equivariance) and linear probe classification (invariance). All models are conditioned on both rotation and color.}  \label{tab:3db_rot_color}
    \centering
\resizebox{0.9\linewidth}{!}{
\centering
     \begin{tabular}{lccc}
        \toprule
        Method & Classification (top-1) & Rotation pred. ($R^2$) & Color pred. ($R^2$) \\
         \midrule
         \midrule
         SEN & 82.17 & 0.29 & 0.96 \\
         EquiMod & 82.58 & 0.27 & 0.95 \\
         Conditional BYOL & 81.95 & 0.38 & 0.94 \\
         SIE & 75.34 & 0.46 & 0.97 \\
         \rowcolor{Light}
         seq-JEPA ($M_{tr}=4$, $M_{val}=5$)  & 79.31 & 0.52& 0.97  \\
         \bottomrule
    \end{tabular}
    }
\end{table}

\subsection{Results with a transformer projector instead of an MLP projector for baselines} \label{sec:projresults}

By using a transformer projector for baselines that normally use an MLP projector, we tested whether transformer-based projectors alone account for any performance gains. We did not see any benefit from switching to transformer projectors in any of the baselines. Table~\ref{tab:trans_projector} shows the transformer-projector results for the baselines on 3DIEBench (plus sign indicates a transformer projector).

\begin{table}[h!]
\centering
\caption{Comparison of baselines trained with transformer projectors on 3DIEBench.}
\label{tab:trans_projector}
\renewcommand{\arraystretch}{1.1}
\begin{tabular}{lccc}
\toprule
Setting\textbackslash Metric & Top-1 Acc & $R^2$ (rel rot) & $R^2$ (abs rot) \\
\midrule
SimCLR+        & 77.92 & 0.32 & 0.48 \\
BYOL+          & 81.05 & 0.09 & 0.16 \\
VICReg+        & 75.10 & 0.15 & 0.28 \\
VICRegTraj+    & 77.82 & 0.24 & 0.36 \\
SEN+           & 81.74 & 0.31 & 0.44 \\
Cond. BYOL+    & 73.62 & 0.27 & 0.36 \\
\bottomrule
\end{tabular}
\end{table}

\subsection{Out-of-distribution results} \label{sec:ood3db}
We created an OOD set for 3DIEBench to enable evaluation on unseen transformations. The original dataset samples rotations from $(-\pi/2, \pi/2)$. Our OOD test set instead uses the disjoint range $(-\pi, -\pi/2) \cup (\pi/2, \pi)$, ensuring no angular overlap with the training set. Rendering this dataset took around five hours on a single A6000 GPU. From Table~\ref{tab:ood_rot}, we see that all methods fail at OOD rotation decoding (with $R^2$s even reaching near $-1.0$). The sharp drop in $R^2$ values may be attributed to a domain shift in transformation geometry: while representations are well-aligned with in-distribution rotation trajectories (inside the range $(-\pi/2, \pi/2)$), they are not geometrically structured to extrapolate beyond this range to $(-\pi, -\pi/2) \cup (\pi/2, \pi)$. Despite the failure in OOD equivariance generalization, seq-JEPA exhibits a graceful degradation in classification accuracy compared to other baselines. We hypothesize that this robustness stems from its sequence-level aggregation mechanism: even when encoder representations become less equivariant under OOD transformations, aggregation across multiple views still filters out transformation-specific variability and recovers shared semantic content. Thus, seq-JEPA maintains object identity better under distribution shift, indicating that aggregation over input views supports invariance-demanding downstream tasks, even when equivariance is imperfect.

\begin{table}[h!]
\centering
\caption{Unseen Transformation Generalization (OOD Rotations)}
\label{tab:ood_rot}
\renewcommand{\arraystretch}{1.1}
\begin{tabular}{lccc}
\toprule
Setting\textbackslash Metric & Top-1 Acc (drop) & OOD $R^2$ (rel rot) & OOD $R^2$ (abs rot) \\
\midrule
SimCLR        & 63.86 (-17.27) & -0.41 & -0.201 \\
BYOL          & 55.68 (-27.22) & -0.25 & -0.198 \\
VICReg        & 61.28 (-19.20) & -0.31 & -0.206 \\
SEN           & 60.03 (-23.40) & -0.41 & -0.201 \\
SIE           & 60.19 (-17.30) & -0.63 & -0.200 \\
EquiMoD       & 58.54 (-25.75) & -0.45 & -0.206 \\
Cond. BYOL    & 57.91 (-24.70) & -0.42 & -0.202 \\
VICRegTraj    & 62.94 (-18.32) & -0.36 & -0.214 \\
seq-JEPA (1,1)& 61.53 (-22.55) & -0.69 & -0.199 \\
seq-JEPA (3,3)& 65.03 (-21.11) & -0.71 & -0.211 \\
\bottomrule
\end{tabular}
\end{table}

\subsection{Equivariance predictor evaluation metrics}

Following the protocol in~\citet{garrido_self-supervised_2023}, we compute and report MRR, Hit@1, and Hit@5 metrics to evaluate predictor quality on the 3DIEBench validation set for seq-JEPA and two predictor-based baselines (Table~\ref{tab:eq_metrics}). These results confirm that seq-JEPA achieves strong equivariance performance not only in terms of $R^2$, but also in top-rank retrieval metrics.

\begin{table}[h!]
\centering
\caption{Predictor evaluation metrics on 3DIEBench validation set.}
\label{tab:eq_metrics}
\renewcommand{\arraystretch}{1.1}
\begin{tabular}{lccc}
\toprule
Setting & MRR & H@1 & H@5 \\
\midrule
SIE              & 0.319 & 0.215 & 0.404 \\
EquiMod          & 0.136 & 0.037 & 0.186 \\
seq-JEPA (1,1)   & 0.340 & 0.241 & 0.442 \\
seq-JEPA (3,3)   & 0.388 & 0.273 & 0.468 \\
\bottomrule
\end{tabular}
\end{table}

\subsection{Training with more compute}

To examine convergence under a high-compute regime, we trained five models using 2000 epochs, $256\times256$ resolution, and batch size 1024. Each of these experiments was run on 4~A100~GPUs (our main experiments were run on a single~A100). The evaluation results (Table~\ref{tab:high_compute}) confirm that seq-JEPA achieves a strong performance in the high-compute regime without suffering from a trade-off between invariance and equivariance. Importantly, our method achieves near-saturated performance already in the low-compute regime, indicating that it requires fewer steps for convergence and is less sensitive to input resolution and batch size than competing methods.

\begin{table}[h!]
\centering
\caption{Evaluation results under the high-compute regime.}
\label{tab:high_compute}
\renewcommand{\arraystretch}{1.1}
\begin{tabular}{lcccccc}
\toprule
Setting & Top-1 Acc & $R^2$ (rel rot) & $R^2$ (abs rot) & MRR & H@1 & H@5 \\
\midrule
SIE              & 82.652 & 0.721 & 0.764 & 0.411 & 0.287 & 0.490 \\
SimCLR           & 85.961 & 0.473 & 0.609 & --    & --    & --    \\
EquiMod          & 86.833 & 0.492 & 0.625 & 0.154 & 0.048 & 0.201 \\
seq-JEPA (1,1)   & 85.370 & 0.661 & 0.713 & 0.365 & 0.263 & 0.447 \\
seq-JEPA (3,3)   & 87.581 & 0.736 & 0.781 & 0.419 & 0.282 & 0.483 \\
\bottomrule
\end{tabular}
\end{table}

\subsection{Complete evaluation results for linear probing on top of aggregate representations} \label{sec:heatmaps}

\begin{figure}[!h]
    \centering
    \includegraphics[width=0.32\linewidth]{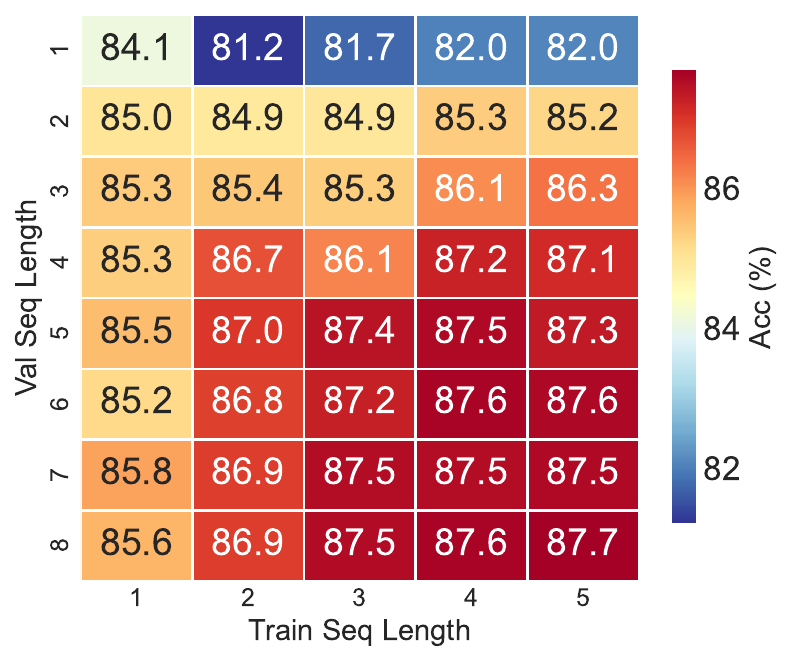}
    \caption{seq-JEPA's performance on 3DIEBench with rotation conditioning; the heatmap shows linear probe accuracy on top of aggregate representations for different training and inference sequence lengths.}
    \label{fig:3db-heatmap}
\end{figure}

\begin{figure}[!h]
    \centering
    \includegraphics[width=0.32\linewidth]{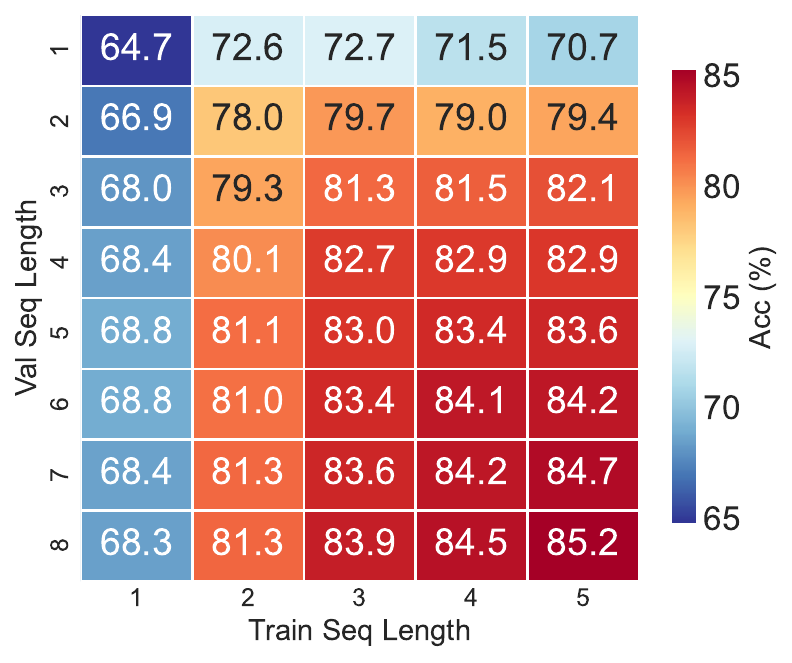}
    \caption{seq-JEPA's performance on STL-10 with predictive learning across saccades; the heatmap shows linear probe accuracy on top of aggregate representations for different training and inference sequence lengths.}
    \label{fig:pls-heatmap}
\end{figure}

\begin{figure}[!h]
    \centering
    \includegraphics[width=0.99\linewidth]{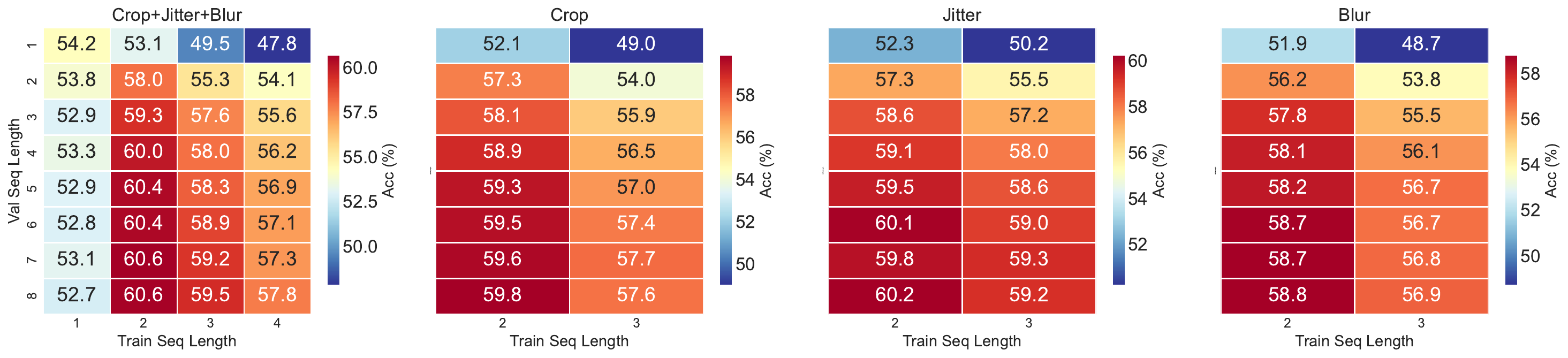}
    \caption{seq-JEPA's performance on CIFAR100 with different types of action conditioning (crop, color jitter, blur, or all three); the heatmap shows linear probe accuracy on top of aggregate representations for different training and inference sequence lengths.}
    \label{fig:cifar100-heatmap}
\end{figure}

\begin{figure}[!h]
    \centering
    \includegraphics[width=0.99\linewidth]{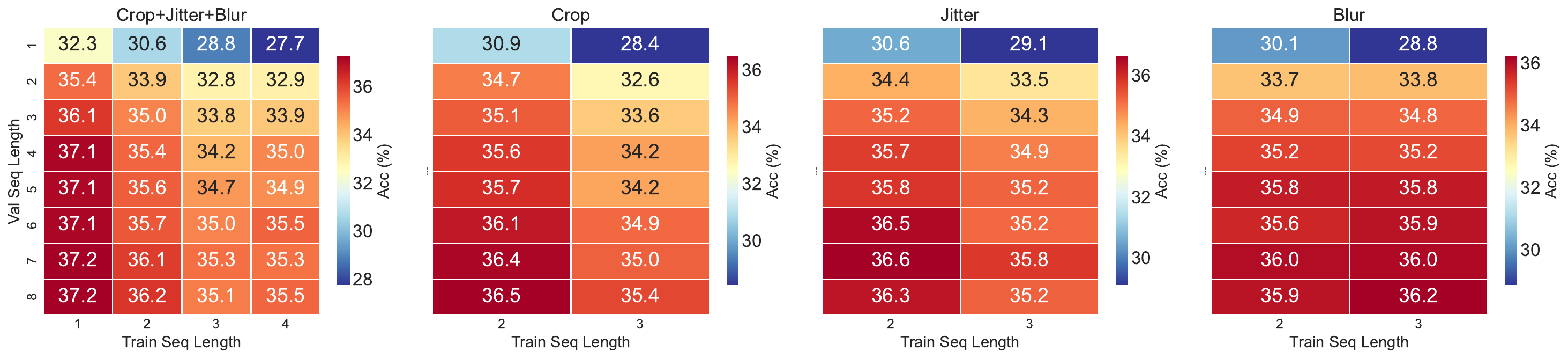}
    \caption{seq-JEPA's performance on Tiny ImageNet with different types of action conditioning (crop, color jitter, blur, or all three); the heatmap shows linear probe accuracy on top of aggregate representations for different training and inference sequence lengths.}
    \label{fig:tinyimg-heatmap}
\end{figure}

We provide our complete evaluation results for linear probing on top of seq-JEPA's aggregate representations for our three transformation settings with different training and inference sequence lengths. Figure~\ref{fig:3db-heatmap} shows the top-1 accuracy on 3DIEBench models conditioned on rotation. Figure~\ref{fig:pls-heatmap} shows top-1 accuracy on STL-10 for models trained via predictive learning across saccades. Figures~\ref{fig:cifar100-heatmap} and \ref{fig:tinyimg-heatmap} show top-1 accuracy on CIFAR100 and Tiny ImageNet, respectively, with different types of action conditioning (crop, color jitter, blur, or all three). These heatmaps reflect the same trends observed in Figure~\ref{fig:seq-len} and discussed in Section~\ref{sec:seqlen}, illustrating the consistent effect of sequence length on representation quality.

\subsection{Comparison of evaluation results on encoder representations and aggregate representations} \label{sec:resaccs}

For completeness, we provide linear probe classification on encoder representations for different transformation settings in Table~\ref{tab:resvsagg} and compare them with accuracy on aggregate representations for different inference evaluation lengths. The aggregate representation generally achieves a much higher classification performance thanks to the architectural inductive bias in seq-JEPA.

\begin{table*}[!h]
\caption{\small Comparison of seq-JEPA classification performance across datasets and conditioning. Top-1 classification accuracy is reported for $z_{res}$ and $z_{agg}$, with varying inference lengths $M_{eval}$.}
\label{tab:resvsagg}
\centering
\resizebox{0.9\linewidth}{!}{
\begin{tabular}{l l c | c c c c}
\toprule
\textbf{Dataset} & \textbf{Conditioning} & \textbf{$M_{tr}$} & \multicolumn{1}{c}{\textbf{$z_{res}$}} & \multicolumn{3}{c}{\textbf{$z_{agg}$}} \\
\cmidrule(lr){4-4} \cmidrule(lr){5-7}
& & & & \textbf{$M_{eval}=1$} & \textbf{$M_{eval}=3$} & \textbf{$M_{eval}=5$} \\
\midrule
3DIEBench & None & 3 & 80.91 & 81.61 & 86.05 & 87.36 \\
3DIEBench & Rotation & 1 & 84.88 & 84.08 & 85.34 & 85.31 \\
3DIEBench & Rotation & 3 & 82.49 & 81.72 & 85.32 & 87.41 \\
3DIEBench & Rotation + Color & 4 & 74.88 & 71.14 & 75.97 & 79.31 \\
\midrule
CIFAR100 & None & 2 & 53.00 & 51.60 & 57.05 & 58.37 \\
CIFAR100 & Crop + Jitter + Blur & 1 & 56.23 & 54.24 & 52.90 & 52.92 \\
CIFAR100 & Crop + Jitter + Blur & 2 & 52.07 & 53.07 & 59.34 & 60.35 \\
CIFAR100 & Crop + Jitter + Blur & 3 & 46.31 & 49.48 & 57.60 & 58.33 \\
CIFAR100 & Crop & 2 & 52.62 & 52.06 & 58.07 & 59.32 \\
CIFAR100 & Color Jitter & 3 & 54.92 & 50.21 & 57.20 & 58.62 \\
CIFAR100 & Blur & 3 & 51.41 & 48.69 & 55.54 & 56.72 \\
\midrule
Tiny ImageNet & None & 2 & 32.84 & 30.48 & 35.03 & 35.97 \\
Tiny ImageNet & Crop + Jitter + Blur & 1 & 33.03 & 32.34 & 36.14 & 37.07 \\
Tiny ImageNet & Crop + Jitter + Blur & 2 & 27.20 & 30.57 & 34.99 & 35.56 \\
Tiny ImageNet & Crop + Jitter + Blur & 3 & 24.74 & 28.84 & 33.78 & 34.68 \\
Tiny ImageNet & Crop & 2 & 31.13 & 30.89 & 35.07 & 35.69 \\
Tiny ImageNet & Color Jitter & 3 & 31.85 & 29.05 & 34.27 & 35.21 \\
Tiny ImageNet & Blur & 3 & 27.20 & 28.83 & 34.82 & 35.79 \\
\midrule
STL-10 & None & 4 & 61.38 & 62.21 & 69.06 & 70.45 \\
STL-10 & Position & 4 & 81.20 & 71.45 & 81.53 & 83.44 \\
STL-10 & Position (no saliency) & 4 & 79.29 & 63.14 & 76.93 & 79.85 \\
STL-10 & Position (no IoR) & 4 & 72.49 & 68.95 & 76.84 & 77.97 \\
\bottomrule
\end{tabular}
}
\end{table*}

\subsection{Additional UMAP Visualizations}
\label{sec:umap}

\begin{figure*}[h]
  \centering
  \includegraphics[width=0.9\linewidth]{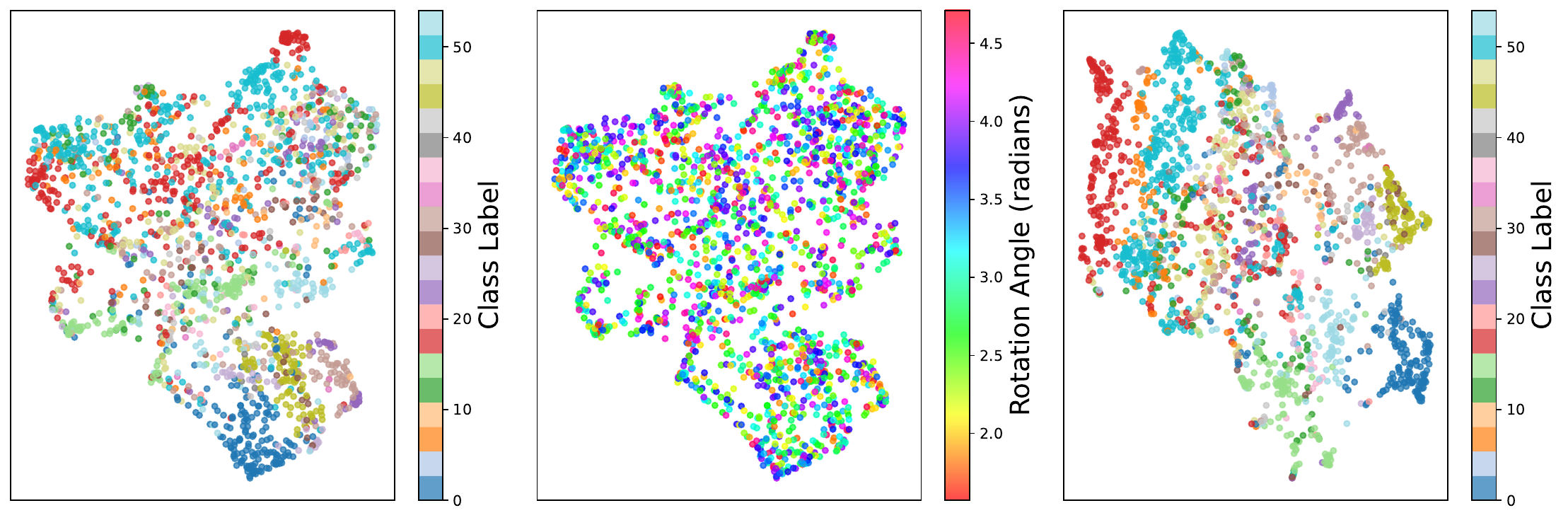}
  \caption{\small 2-D UMAP projections of seq-JEPA representations on 3DIEBench \emph{without} action conditioning ($M_{tr}=3$, $M_{val}=5$). Encoder outputs are color-coded by class (\textbf{left}) and rotation angle (\textbf{middle}); aggregate token representations are color-coded by class (\textbf{right}).}
  \label{fig:umap-3db-noact}
\end{figure*}

\begin{figure*}[h]
  \centering
  \includegraphics[width=0.6\linewidth]{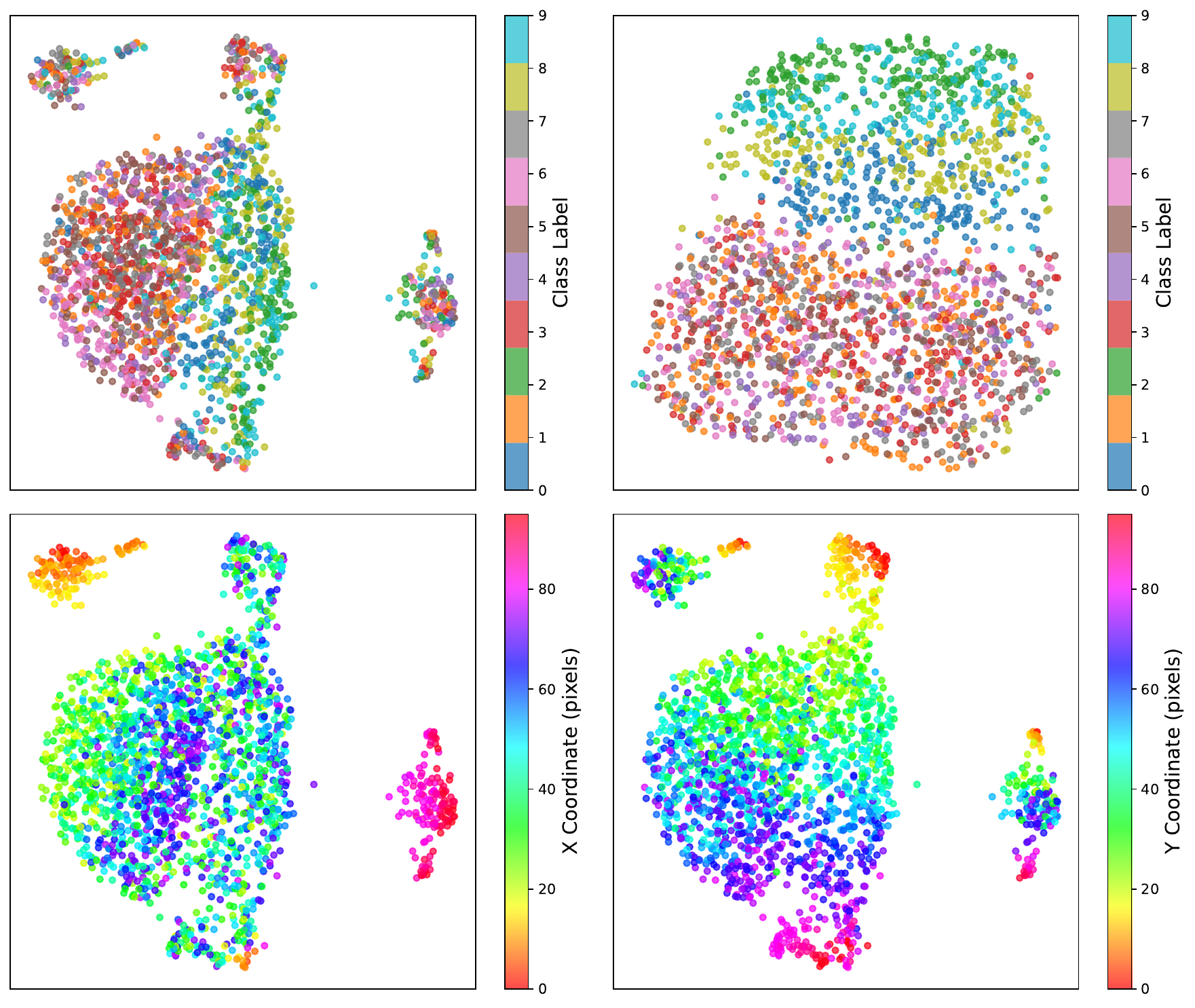}
  \caption{\small 2-D UMAP projections of seq-JEPA representations on STL-10 with action conditioning ($M_{tr}=M_{val}=4$). Encoder outputs are color-coded by class (top-left) and by $X$/$Y$ fixation coordinates (bottom); the aggregate token is color-coded by class (top-right).}
  \label{fig:umap-pls}
\end{figure*}

\begin{figure*}[h]
  \centering
  \includegraphics[width=0.9\linewidth]{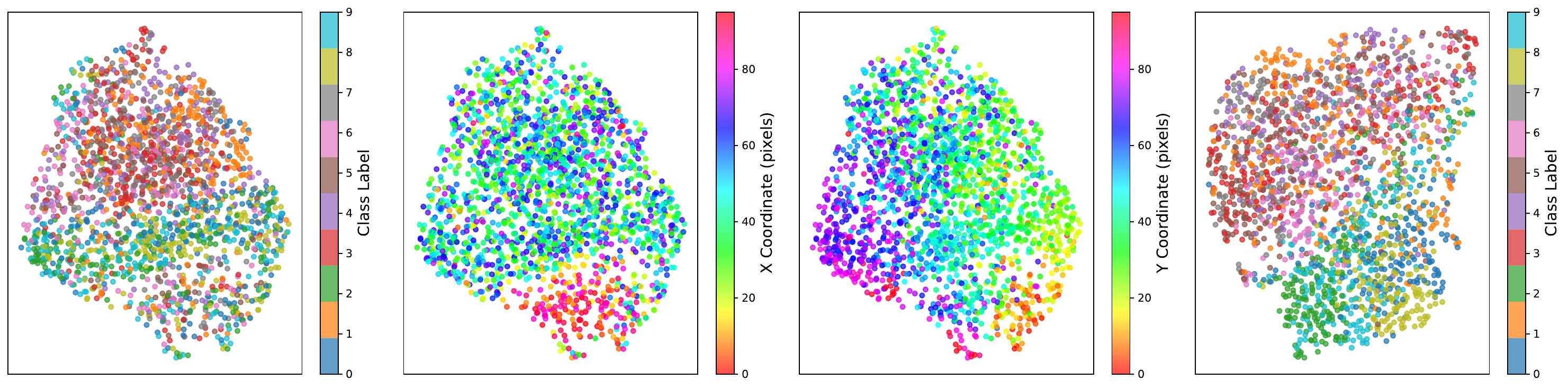}
  \caption{\small 2-D UMAP projections of seq-JEPA representations on STL-10 \emph{without} action conditioning ($M_{tr}=M_{val}=4$). Encoder outputs are color-coded by class (\textbf{left}) and fixation coordinates (\textbf{middle}); the aggregate token is color-coded by class (\textbf{right}).}
  \label{fig:umap-pls-noact}
\end{figure*}

In Figure~\ref{fig:umap-3db-noact}, we visualize the UMAP projections of seq-JEPA representations trained on 3DIEBench \emph{without} action conditioning. Similarly, Figure~\ref{fig:umap-pls-noact} shows projections for models trained on STL-10 via predictive learning across saccades, also without action conditioning. Compared to the action-conditioned counterparts (Figures~\ref{fig:umap-3db} and~\ref{fig:umap-pls}), these projections exhibit weaker or no smooth color gradients in their corresponding transformation-colored UMAP—indicating reduced equivariance to transformation parameters.

\subsection{Details of Path Integration Experiments.} \label{sec:pathint}
While an agent executes a sequence of actions in an environment, transitioning from an initial state to a final state, it should be capable of tracking its position by integrating its own actions. This is also a crucial cognitive ability that enables animals to estimate their current state in their habitat \citep{mcnaughton_path_2006}. Here, we evaluate whether seq-JEPA is capable of \emph{path integration}. Given the sequence of observations $\{x_i\}_{i=1}^{M+1}$ generated from transformations $\{t_i\}_{i=1}^{M+1}$ and the corresponding relative action embeddings $\{a_i\}_{i=1}^{M}$, we define the task of path integration over the sequence of actions as predicting the relative action that would directly transform $x_1$ to $x_{M+1}$ given $z_{AGG}$ and $a_{M}$. In other words, given the aggregate representation of a sequence of action-observation pairs and the next action, we would like to predict the overall position change from the starting point ($x_1$) to the end point ($x_{M+1}$). We consider path integration for rotation angles in 3DIEBench and across eye movements with STL-10. For rotations, the task is integrating a series of object rotations from the first view to the last, i.e. angular path integration. For eye movements, the task is integrating the eye movements from the first saccade to the last, i.e. visual path integration. To measure path integration performance for inference sequence length $M$, we train a regression head on top of the concatenation of $z_{AGG}$ and $a_M$ to predict the transformation from $x_1$ to $x_{M+1}$. Figure~\ref{fig:path-int} shows that seq-JEPA performs well in both angular and visual path integration. The red curve corresponds to the performance of the original seq-JEPA. The blue curve corresponds to experiments in which the action embeddings are ablated (zeroed-out during inference for all views). The green curve corresponds to experiments in which the encoder (visual) representations are ablated during inference. As expected, path integration becomes more difficult as the number of observations increases (red curves). Ablating action conditioning (blue curves) results in failure of path integration. On the other hand, ablating the visual representations (green curves) results only in a small performance drop compared to the original model, indicating that action conditioning is the key factor that enables path integration.

\subsection{Transfer learning results on ImageNet-1k} \label{sec:transfer}

To evaluate generalization of our model beyond STL-10, we assess transfer performance of the model trained on STL-10 patches via predictive learning across saccades on ImageNet-1k. We follow the same linear probing protocol as in-distribution evaluations: we freeze the ResNet and transformer encoder and train a linear classifier on aggregate representations generated from foveated patches.

\begin{figure}[H]
    \centering
    \includegraphics[width=0.3\linewidth]{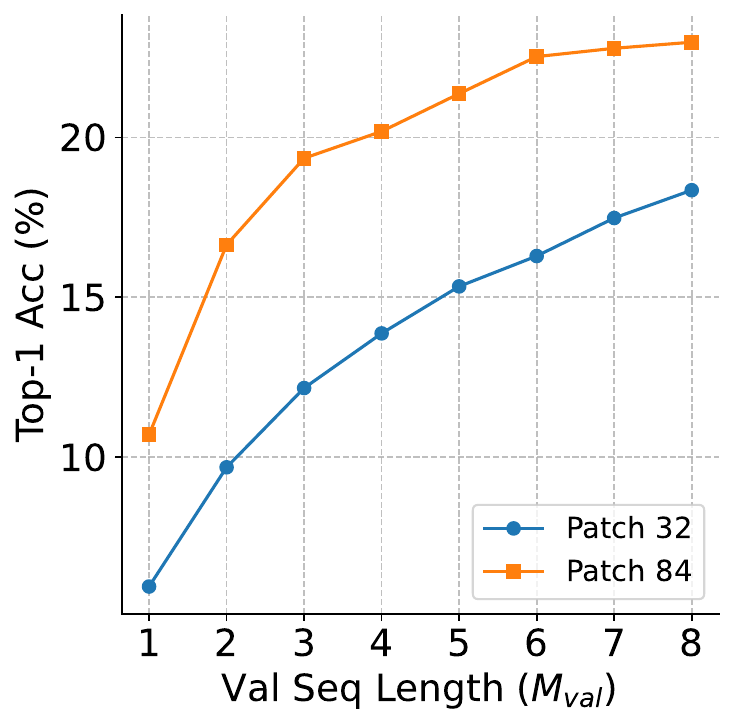}
    \caption{Linear probe transfer learning accuracy on ImageNet-1k for two different patch sizes; the model is pre-trained on STL-10 via predictive learning across saccades.}
    \label{fig:transfer-imgnet}
\end{figure}

We extract saliency maps for ImageNet-1k images. Saliencies were extracted using DeepGaze IIE with the MIT1003 centerbias prior~\citep{linardos_deepgaze_2021}. Maps are saved at native resolution (matching the original ImageNet image dimensions) and normalized to probability distributions. For our transfer learning experiments we resize the images and saliencies to $224\times224$, and sample sequences of patches sized $32\times32$ or $84\times84$ for this evaluation setting. Figure~\ref{fig:transfer-imgnet} shows top-1 linear probe accuracy on ImageNet-1k validation set across varying inference sequence lengths ($M_{val}$). For both patch sizes, performance improves with longer inference sequences, validating seq-JEPA’s ability to benefit from extended context even in this difficult OOD ImageNet-1k setting. These results echo findings in the main experiments for different training and inference sequence lengths and confirming the possibility of model's scalability in terms of data, parameter count, and compute.

\end{document}